\documentclass[letterpaper]{article} 
\usepackage{aaai24}  
\usepackage{times}  
\usepackage{helvet}  
\usepackage{courier}  
\usepackage[hyphens]{url}  
\usepackage{graphicx} 
\usepackage{natbib}  
\usepackage{caption} 
\usepackage{algorithm}
\usepackage{algorithmic}

\usepackage{newfloat}
\usepackage{listings}
\DeclareCaptionStyle{ruled}{labelfont=normalfont,labelsep=colon,strut=off} 
\floatstyle{ruled}
\newfloat{listing}{tb}{lst}{}
\floatname{listing}{Listing}
\usepackage{multirow}
\usepackage{booktabs}
\usepackage{xspace}
\usepackage{xcolor}
\usepackage{amsmath}
\usepackage{arydshln}
\usepackage{amsfonts}
\usepackage{pgfplotstable}
\usepackage{subfigure} 
\usepackage{tikz,pgfplots}
\pgfplotsset{compat=1.18}

\newcommand\scalemath[2]{\scalebox{#1}{\mbox{\ensuremath{\displaystyle #2}}}}
\definecolor{ForestGreen}{RGB}{34,139,34}
\definecolor{BrickRed}{rgb}{0.8, 0.25, 0.33}
\definecolor{Plum}{rgb}{0.56, 0.27, 0.52}
\definecolor{Cerulean}{rgb}{0.0, 0.48, 0.65}
\definecolor{RoyalPurple}{rgb}{0.47, 0.32, 0.66}
\definecolor{Dandelion}{rgb}{0.94, 0.88, 0.19}
\definecolor{Red}{rgb}{1.0, 0.0, 0.0}
\definecolor{Blue}{rgb}{0.0, 0.0, 1.0}
\definecolor{Azure}{rgb}{0.0, 0.5, 1.0}
\definecolor{Golden}{rgb}{1.0, 0.84, 0.0}
\definecolor{blueaccent}{RGB}{178,223,242}
\definecolor{greenaccent}{RGB}{0,139,43}
\definecolor{purpleaccent}{RGB}{130,41,128}
\definecolor{orangeaccent}{RGB}{242,197,178}

\newcommand*{\eg}{\emph{e.g.}\@\xspace}
\newcommand*{\ie}{\emph{i.e.}\@\xspace}
\newcommand*{\etc}{\emph{etc.}\@\xspace}
\newcommand*{\vs}{\emph{vs.}\@\xspace}

\newcommand*{\elbafont}{\fontfamily{augie}\selectfont}
\newcommand*\modelname{{\scriptsize \elbafont CORONET}\xspace}

\newcommand*{\query}{{\(Q\)}}

\title{Commonsense for Zero-Shot Natural Language Video Localization}
\author {
    Meghana Holla\textsuperscript{\rm 1},
    Ismini Lourentzou\textsuperscript{\rm 2},
}
\affiliations {
    \textsuperscript{\rm 1}Department of Computer Science, Virginia Tech\\
    \textsuperscript{\rm 2}School of Information Sciences, University of Illinois at Urbana - Champaign\\
    mmeghana@vt.edu,
    lourent2@illinois.edu
}

\begin{document}

\maketitle

\begin{abstract}
Zero-shot Natural Language-Video Localization (NLVL) methods have exhibited promising results in training NLVL models exclusively with raw video data by dynamically generating video segments and pseudo-query annotations.
However, existing pseudo-queries often lack grounding in the source video, resulting in unstructured and disjointed content. In this paper, we investigate the effectiveness of commonsense reasoning in zero-shot NLVL. Specifically, we present \modelname, a zero-shot NLVL framework that leverages commonsense to bridge the gap between videos and generated pseudo-queries via a commonsense enhancement module. \modelname employs Graph Convolution Networks (GCN) to encode commonsense information extracted from a knowledge graph, conditioned on the video, and cross-attention mechanisms to enhance the encoded video and pseudo-query representations prior to localization. Through empirical evaluations on two benchmark datasets, we demonstrate that \modelname surpasses both zero-shot and weakly supervised baselines, achieving improvements up to $32.13\%$ across various recall thresholds and up to $6.33\%$ in mIoU. These results underscore the significance of leveraging commonsense reasoning for zero-shot NLVL.
\end{abstract}

\section{Introduction}
\label{sec:intro}

Natural Language Video Localization (NLVL) is a fundamental multimodal understanding task that aims to align textual queries with relevant video segments. NLVL is a core component for various applications such as video moment retrieval \cite{cao_visual_2022}, video question answering \cite{qian_locate_2022,lei_tvqa_2020}, and video editing \cite{gao_end--end_2022}. Prior works have primarily explored supervised 
\cite{zeng_dense_2020, 
wang_temporally_2020, 
soldan_vlg-net_2021, 
liu_context-aware_2021, 
yu_intra-_2020, 
gao_relation-aware_2021} 
or weakly supervised \cite{mun_local-global_2020, zhang_counterfactual_2020, zhang_video_2021} NLVL methodologies, relying on annotated video-query data to various extents.

\begin{figure}[t!]
    \centering
    \includegraphics[width=0.9\linewidth]{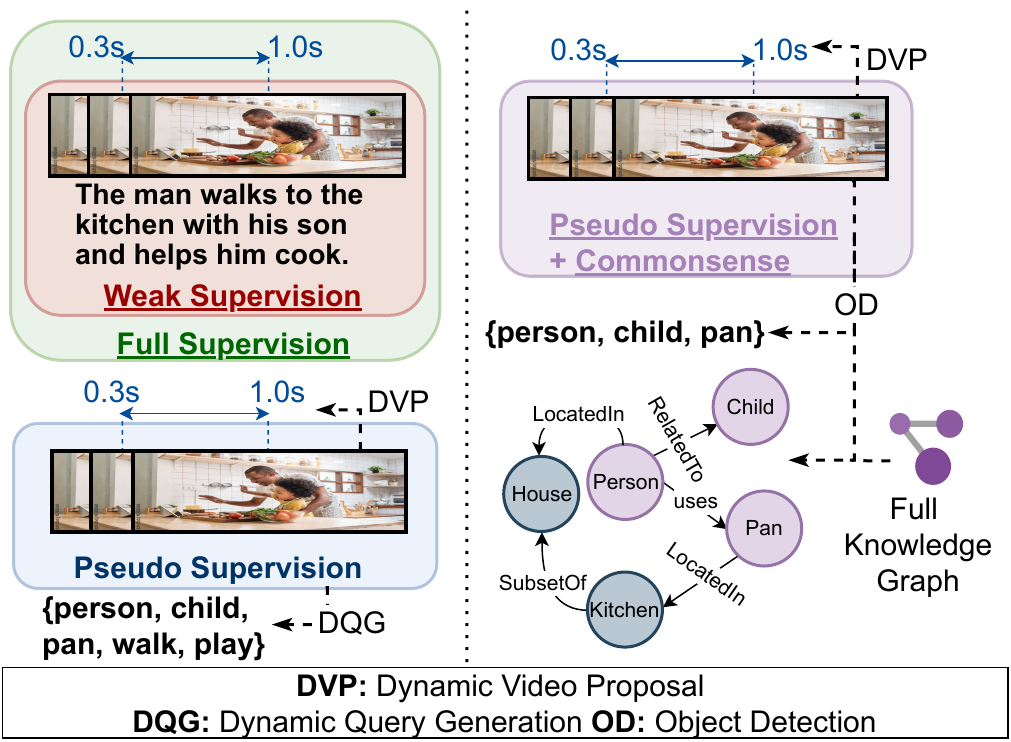}
    \caption{NLVL tasks under various supervision settings.  Color-coded boxes show the expected annotations at each supervision level. \textbf{\textcolor{ForestGreen}{Full supervision}}: Temporal Video Annotations + Text Queries; \textbf{\textcolor{BrickRed}{Weak Supervision}}: Text Queries; \textbf{\textcolor{Azure}{Pseudo-Supervision}}: Only Raw Videos. DVP + DQG; \textbf{\textcolor{Plum}{\modelname~(Ours, right)}} Only Raw Videos. DVP + OD and video-informed commonsense knowledge subgraph.}
    \label{fig:teaser}
    
\end{figure}

Obtaining annotated data for NLVL is a labor-intensive process that requires video samples paired with meticulous annotations of video moments and corresponding textual descriptions. Figure \ref{fig:teaser} illustrates the annotation requirements for different levels of supervision in NLVL. Fully supervised methods demand fine-grained moment span annotations, while weakly supervised methods typically rely on query descriptions alone. Nevertheless, both still heavily rely on paired video-language data, which limits practicality in open-domain settings.

Recent works formulate zero-shot NLVL, which aims to dynamically generate video moments and their corresponding queries, eliminating the need for paired video-query data \cite{nam_zero-shot_2021,kim2023language}. Nonetheless, existing approaches have certain limitations. On one hand, recent methods generate pseudo-queries using off-the-shelf object detectors for objects (nouns) and text-based language models for actions (verbs), resulting in noisy pseudo-queries that lack grounding in the video content \cite{nam_zero-shot_2021}. On the other hand, language-free methods remove pseudo-queries entirely by utilizing vision-language models pretrained on large-scale image and text datasets \cite{kim2023language}. However, eliminating textual information entirely may lead to missing out on important semantic nuances.

Visual (video) and textual (query) modalities provide very distinct but complementary types of information; videos provide spatial and physical information, while queries provide situational and contextual information. Existing works focus on complex vision-language interactions for observed video-query pairs in an attempt to bridge this gap~\cite{nam_zero-shot_2021,mun_local-global_2020}. However, in the zero-shot/pseudo-supervised setting, where queries are in a simpler form without structural information, finding common ground between modalities becomes crucial for effective cross-modal interactions. Commonsense knowledge, which encompasses general knowledge about the world and relationships between concepts, has proven valuable in various tasks \cite{fang_video2commonsense_2020,ding_dynamic_2021,yu_hybrid_2021,li_representation_2022,maharana_integrating_2021,cao_visual_2022}. By incorporating commonsense information, NLVL models could potentially bridge the semantic gap between video and text modalities, enhancing the cross-modal understanding and performance in zero-shot NLVL.

To this end, this work introduces  \textbf{C}omm\textbf{O}nsense ze\textbf{R}o sh\textbf{O}t la\textbf{N}guage vid\textbf{E}o localiza\textbf{T}ion (\textbf{\modelname}), a zero-shot NLVL model that leverages commonsense knowledge to enhance the pseudo-query generation and cross-modal localization of video moments. We introduce a Commonsense Enhancement Module to enrich the encoded video and query representations with rich contextual information and employ external commonsense knowledge from ConceptNet~\cite{speer_conceptnet_2017} to extract relevant relationships between a predefined set of concepts, mined from the input videos. Our primary objective is to investigate the potential benefits and challenges of leveraging commonsense for zero-shot NLVL. By jointly incorporating commonsense knowledge, we show that our model effectively bridges the gap between visual and linguistic modalities.

The contributions of this work are summarized as follows:
\noindent \textbf{(1)} We introduce \modelname\footnote{Code available at \url{https://github.com/PLAN-Lab/CORONET}}, a zero-shot NLVL framework that utilizes external commonsense knowledge to enrich cross-modal understanding between the visual and natural language components of pseudo-query generation. To the best of our knowledge, we are the first to incorporate commonsense information in zero-shot natural language video localization. 
\noindent \textbf{(2)} \modelname extracts knowledge subgraphs that can be employed to enrich vision-language understanding effectively and an accompanying commonsense enrichment module that can be easily integrated into video localization. 
\noindent \textbf{(3)} We provide empirical evidence of the effectiveness of our approach, demonstrating improvements up to $32.13\%$ across various recall thresholds and up to $6.33\%$ in mIoU. Extensive ablation studies thoroughly investigate the impact of commonsense on zero-shot NLVL performance. 
\section{Related Work}
\label{relatedWorks}
\subsection{Natural Language Video Localization (NLVL)}
Previous works on NLVL can be categorized into proposal-based~\cite{liu_exploring_2022, gao_relation-aware_2021, soldan_vlg-net_2021, xiao_boundary_2021, yang_deconfounded_2021, gao_fast_2021, yu_intra-_2020, wu_learning_2022} and proposal-free approaches~\cite{rodriguez_proposal-free_2020, chen_hierarchical_2020, mun_local-global_2020, zeng_multi-modal_2021, zhao_cascaded_2021, zhang_natural_2022, rodriguez-opazo_dori_2021}. Proposal-based methods employ a generate-and-rank strategy, \ie generating candidate video moments and subsequently ranking them based on their alignment with the given textual query. In contrast, proposal-free methods directly regress on the untrimmed video, estimating the boundaries of the target video segment based on the query.

The majority of NLVL works are fully supervised, with proposal-free methods primarily focusing on segment localization or regression accuracy~\cite{zeng_dense_2020, wang_temporally_2020, rodriguez-opazo_dori_2021}, while proposal-based concentrating on improving the quality of the proposed video moment candidates~\cite{xiao_boundary_2021}. To effectively capture cross-modal relationships, several works transform either the video or query modalities, or both, into graphs and perform graph matching~\cite{soldan_vlg-net_2021, rodriguez-opazo_dori_2021, zeng_multi-modal_2021, chen_hierarchical_2020}. Some proposal-free works utilize convolutions to capture long-span dependencies within videos~\cite{li_proposal-free_2021} or as a form of cross-modal interaction~\cite{zhang_natural_2022, chen_hierarchical_2020}. Moreover, there exist works that reframe NLVL into a generative task~\cite{li2023momentdiff} or traditional NLP tasks such as multiple-choice reading comprehension~\cite{gao_relation-aware_2021} and dependency parsing~\cite{liu_context-aware_2021}. 

\subsection{Weakly Supervised and Zero-shot NLVL Methods}
Fully supervised methods achieve impressive performance but require laborious
fine-grained video segment annotations corresponding to queries that are often prohibitively expensive for adapting to new domains. To address this challenge, weakly supervised methods have emerged, which operate with paired video-query data but without the need for precise video segment span annotations 
\cite{huang_cross-sentence_2021,zhang_counterfactual_2020,ma_vlanet_2020, 
detr}.
Many weakly supervised approaches leverage contrastive learning to improve visual-textual alignment \cite{zhang_counterfactual_2020, zhang_video_2021, ma_vlanet_2020}. Recent work employs graph-based methodologies to capture contextual relationships between frames \cite{tan_logan_2021} and iterative approaches for fine-grained alignment between individual query tokens and video frames \cite{wang_fine-grained_2021}. 

\begin{figure*}[t!]
    \centering
    \includegraphics[width=0.95\textwidth]{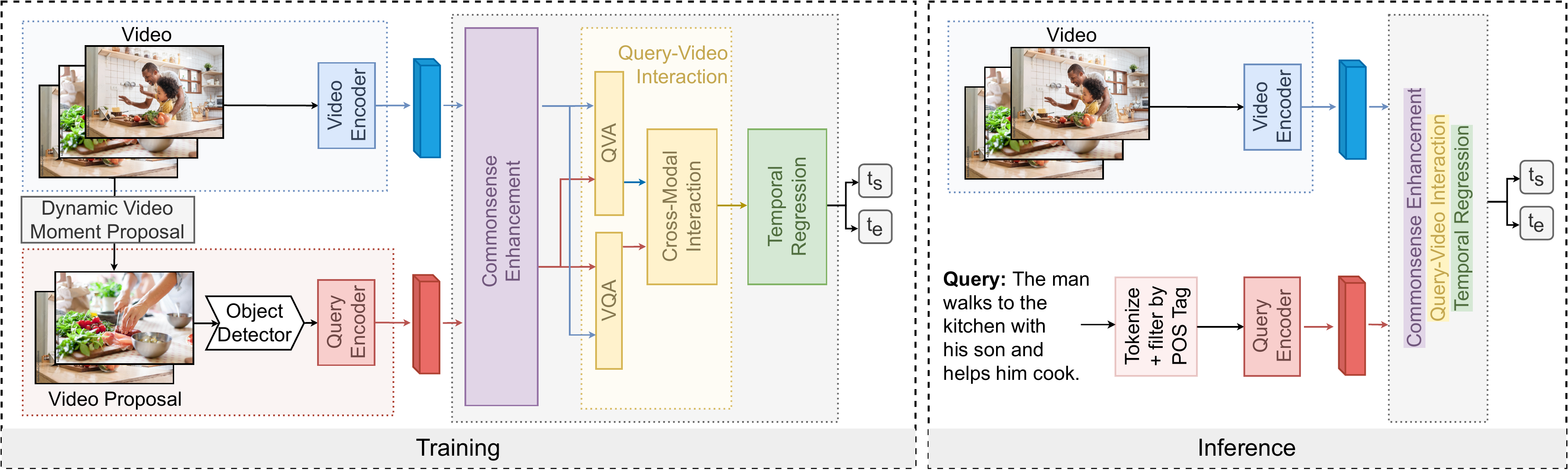}
    \caption{\modelname consists of a {\textcolor{Cerulean}{Video Encoder}} and a {\textcolor{Red}{Query Encoder}}, the proposed {\textcolor{RoyalPurple}{Commonsense Enhancement}}, a {\textcolor{Golden}{Cross-modal (video-query) Interaction}}, and a {\textcolor{ForestGreen}{Temporal Regression}} module. During training, \modelname utilizes a Dynamic Video Moment Proposal module to extract a video moment span \(V_{\text{span}}\) and an off-the-shelf object detector to detect objects (nouns) in \(V_{\text{span}}\). During inference, the given natural language query is converted to a simplified query using a part-of-speech tagger. }
    \label{fig:approach}
\end{figure*}

Despite requiring fewer annotations, the effort involved in acquiring queries is still substantial. Unsupervised iterative approaches \cite{liu_unsupervised_2022} and zero-shot NLVL (ZS-NLVL) \cite{nam_zero-shot_2021} address this issue. ZS-NLVL aims to train an NLVL model using raw videos alone in a self-supervised setting, by generating video moments and corresponding pseudo-queries dynamically.
Pseudo-query generation is critical in zero-shot localization methods, although limited work has been done in this direction. \citet{nam_zero-shot_2021} introduce pseudo-query generation for video localization, and subsequently, \citet{jiang_pseudo-q_2022} for language grounding in images.
\citet{nam_zero-shot_2021} consider a pseudo-query to be an unordered list of nouns and verbs, obtained from an off-the-shelf object detector and a fine-tuned language model (LM) that predicts the most probable verbs conditioned on the nouns. While the objects are grounded in the video segment, the generation of verbs is not, potentially introducing irrelevant verbs and resulting in noisy pseudo-queries. Moreover, explicit verb-noun co-occurrences may encourage the localization model to learn spurious latent relationships and co-occurrence patterns between noun and verb data.
\citet{kim2023language} propose a language-free approach that leverages the aligned visual-language space of a pretrained CLIP model. A limitation is primarily relying on visual and temporal cues for video grounding but not fully capturing higher-level contextual knowledge and implicit relationships often conveyed through natural language. This could hinder the model's ability to understand and localize events in videos when additional context is necessary.
In contrast, \modelname enriches the extracted video and pseudo-query features with commonsensical information. By considering spatiotemporal, causal, and physical relations w.r.t. the visual information, our model reasons beyond video cues and grounds pseudo-query information in the video.

\subsection{Commonsense in Video-Language Tasks}
Recent video-language research has shifted towards enhancing reasoning capabilities rather than solely focusing on recognition. Datasets such as Video2Commonsense \cite{fang_video2commonsense_2020}, Something Something \cite{goyal_something_2017}, Violin \cite{liu_violin_2020}, SUTD-TrafficQA \cite{peng_multi-modal_2021}, and VLEP \cite{lei_what_2020} emphasize commonsense reasoning. Metrics have also been proposed to evaluate the commonsense reasoning abilities of video-language models \cite{shin_cogme_2021,park_exposing_2022}. 
Commonsense has also been incorporated into tasks such as video captioning \cite{yu_hybrid_2021}, video question answering \cite{li_representation_2022}, and visual story generation \cite{maharana_integrating_2021}. 
Existing methods enhance query-based video retrieval using a co-occurrence graph of concepts mined from the target video moment \cite{wu_learning_2022,cao_visual_2022}. However, both are proposal-based fully supervised approaches that rely on fine-grained annotations and the quality of candidate video moments, let alone solely exploit the internal relations between the detected visual objects through a co-occurrence graph of entities as opposed to using external knowledge sources. In contrast, we utilize structured knowledge sources such as ConceptNet \cite{speer_conceptnet_2017} to encode commonsense information and leverage explicit relations spanning spatial, temporal, and physical aspects. This allows us to access information beyond what visual and textual cues can provide.

\section{Commonsense for Zero-Shot NLVL}
\label{sec:proposedSection}

\subsection{Problem Formulation}
We denote an input video as $V$, and its grounding annotations as \(\left( Q,V_{\text{span}}\right) \), where $Q$ is the query representation and \(V_{\text{span}}\!=\!\left( t_{s},t_{e}\right)\) is the corresponding video moment span annotation, with \(t_{s}\) and \(t_{e}\) representing the start and end timestamps, respectively. Learning to localize a video moment conditioned on a query entails maximizing the expected log-likelihood of the model parameterized by \(\theta\). In its typical setting, this can be formulated as follows:
\begin{equation}
\label{eq:groundingOriginal}
    \theta ^{\ast }=\arg \max _{\theta } \mathbb{E}\left[ \log p_{\theta }\left(  V_{\text{span}} | V,Q\right) \right]. 
\end{equation}
In the zero-shot setting, the goal is to learn this task without parallel video-query annotations. Hence, the query and video moment annotations are derived from $V$, using a dynamic video moment proposal method followed by a pseudo-query generation mechanism. Formally,  \(V_{\text{span}}\,\!{=}\!\,f_{\text{span}}(V)\) and \(Q\,\!{=}\!\,f_{pq}(V_{\text{span}})\), where $f_{\text{span}}$ and $f_{\text{pq}}$ are video moment proposal and pseudo-query generation mechanisms, respectively. Given that $f_{\text{span}}$ and $f_{\text{pq}}$ are responsible for generating the annotations, the performance of the localization model heavily depends on the quality of these modules. Existing methods face challenges in aligning \(Q\) to \(V_{\text{span}}\) due to noise introduced by ungrounded pseudo-query generation mechanisms. 
To address this, we simplify \(f_{\text{pq}}\) while augmenting cross-modal understanding by leveraging external information in the form of a commonsense graph \(G_{C}(C, E)\) with \(n_c\) nodes, where \(C\!=\!\left\{c_{1}, c_{2}, \dots, c_{n_{C}}\right\}\) are the concept node vector representations and \(E\) is the set of weighted directed edges, respectively. Accordingly, learning can be formulated as
\begin{equation}
\label{eq:groundingOurs}
    \theta ^{\ast }=\arg \max _{\theta } \mathbb{E}\left[ \log p_{\theta }\left(  V_{\text{span}}| V,Q,G_{C}\right) \right].
\end{equation}

\noindent Figure \ref{fig:approach} shows both training and inference flows.
\subsection{Pseudo-supervised Setup}
\modelname first processes a raw video with a video moment proposal $f_{\text{span}}$ module that extracts important video segments capturing key events, and a pseudo-query generation $f_{\text{pq}}$ that generates text query annotations corresponding to the extracted video segments.

\paragraph{Dynamic Video Moment Proposal ($f_{\text{span}}$).}
We adopt the dynamic video moment proposal approach proposed by \citet{nam_zero-shot_2021}. Specifically, $f_{\text{span}}$ primarily comprises a k-means clustering mechanism that groups semantically similar and temporally proximal video frame features together to extract atomic moments. To obtain frame features, we consider the columns of a frame-wise similarity matrix derived from the CNN features of individual frames. We enforce temporal proximity by concatenating the frame index to the features. Composite video moments are then formed by combining neighboring atomic moments, and a subset of all possible combinations is sampled uniformly at random. The resulting set of video moments corresponds to $V_{\text{span}}$.

\paragraph{Pseudo-query Generation ($f_{\text{pq}}$).} The pseudo-query is constructed as a collection of objects present in the video. To generate the pseudo-query, we employ an off-the-shelf object detector, enabling the extraction of pertinent objects in \(V_{\text{span}}\). We adopt a top-$k$ strategy to sample the $k$ most probable object predictions associated with the query \query.

\paragraph{Video Encoder.}
We uniformly sample $T$ frames from $V$ and extract their CNN (\eg, I3D~\cite{qian_locate_2022}) features. These features are contextually encoded using a video encoder ${\phi}_{v}$ to yield frame features ${\phi}_{v}(V)\!=\!\left\{ v_{1},v_{2},\ldots,v_{T}\right\}$ where $v_{i}\in\mathbb{R}^{d}$, and $d$ is the common video/query encoding dimension. We implement ${\phi}_{v}$ as a GRU-based encoder.

\paragraph{Query Encoder.}
Our pseudo-query $Q$, composed of up to $k$ tokens, is encoded using a query encoder ${\phi}_{q}$ that generates query embeddings ${\phi}_{q}(Q)\!=\!\left\{ q_{1},q_{2},\ldots,q_{k}\right\}$, for the top-$k$ detected objects extracted from the pseudo-query generation. Here, $q_{i}\in \mathbb{R}^{d}$ and $d$ is the common video/query encoding dimension. We implement ${\phi}_{q}$ as a bi-directional GRU-based encoder preceded by a trainable embedding layer. 

\subsection{Commonsense Enhancement Module}
\label{sec:cem}
\begin{figure}[t!]
    \centering
    \includegraphics[width=0.85\linewidth]{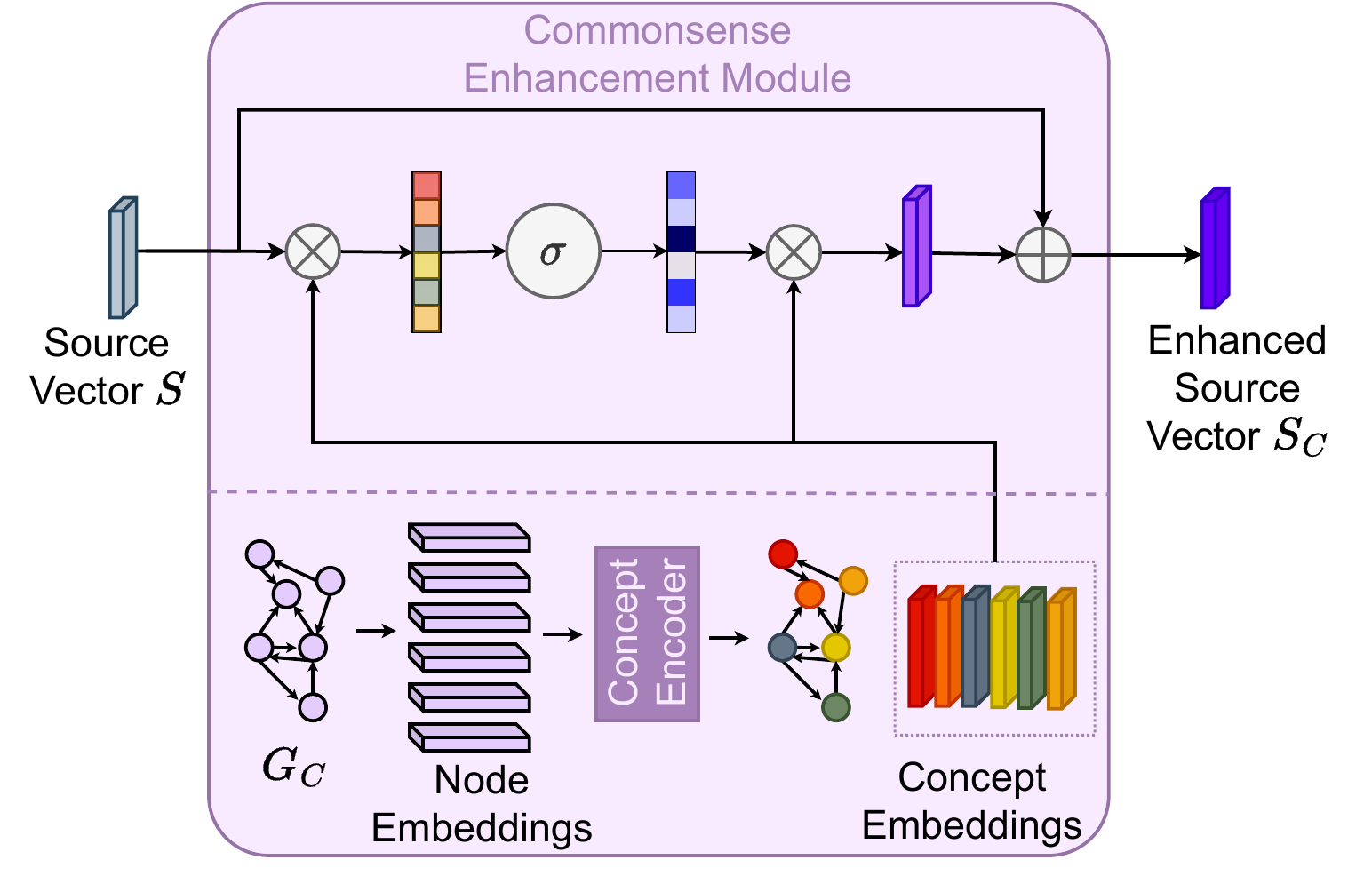}
    \caption{\modelname Commonsense Enhancement Module (CEM). CEM comprises a concept encoder and an enhancement mechanism that uses the previously encoded concept vectors to update a given input vector (video/query vectors). The concept encoder employs a Graph Convolution Network for encoding the nodes (concepts) of \(G_C\). 
    }
  \label{fig:cem}
\end{figure}

To enrich the encoded video and query features with information grounded in commonsensical knowledge, we introduce a Commonsense Enhancement Module (CEM), pictorially described in Figure~\ref{fig:cem}. This enhancement helps inject necessary information into video and query representations, which can not just help bridge the gap between the available visual and textual cues but also provide rich information to the downstream span localization module. 

CEM includes a set \(C\!=\!\left\{c_{1}, c_{2}, \dots, c_{n_{C}}\right\}\) of \(n_{C}\) concept vectors, where \(c_{i} \in \mathbb{R}^{d}\) and \(d\) is the concept feature dimension (same dimension as $\forall v_i \in V$ and $\forall q_i \in Q$). In general, given source feature vectors $S\!=\!\left\{ s_{1},s_{2},\ldots,s_{n}\right\}$ with individual feature vectors $s_{i \in [1,n]} \in \mathbb{R}^{d}$, the enhanced feature vectors $S_{C}$ are obtained using a commonsense enhancement mechanism $\phi_{C}$.
We implement this commonsense enhancement step $\phi_{C}$ as a cross-attention mechanism that enriches source input features, attending over $S$ guided by the commonsense concept vectors $C$, \ie, 
\begin{equation}
\label{eq:cenhance}
\scalemath{1}{
    }
    S_{C} = S + \phi_{C}(S) = S + \sigma \left( \frac{SW_{Q}(CW_{K})^{T}}{\sqrt{d}} \right) C W_{V},
\end{equation}
where $\sigma$ is a softmax activation, \(W_{Q}\), \(W_{K}\), \(W_{V}\) are trainable matrices and \(d\) is the common dimension of the vectors \(S\) and \(C\). In our setting, the source feature vectors $S$ are either video $V$ or pseudo-query $Q$ features. We build separate enhancement mechanisms for $V$ and $Q$, \ie, the projection matrices \(W_{Q}\), \(W_{K}\), \(W_{V}\) are not shared between $Q$ and $V$. We elaborate more on the rationale in the appendix.
The enriched video and pseudo-query features are denoted as \(V_{C}\!=\!\phi_{C_{\text{vid}}}(V)\) and \(Q_{C}\!=\!\phi_{C_{\text{pq}}}(Q)\), respectively.

\paragraph{Concept Encoder.}
The concept vectors \(C\) mentioned above are feature representations that internally form the nodes of the commonsense graph, \(G_C\). Accordingly, graph \(G_{C}\) is represented as a matrix, where \(G_{C(i,j)}\) represents the total number of directed relational edges between \(c_{i},c_{j} \in C\) that start at \(c_i\) and end at \(c_j\). To encode the commonsense information, we employ Graph Convolutional Networks (GCN) \cite{hammond_wavelets_2011}. The concept encoder is composed of $L$ graph convolution layers, each of which performs a convolution step
\begin{equation}
\scalemath{1}{
    C^{\left(l+1\right)}=\sigma \left( AC^{\left(l\right) }W^{\left( l\right) }\right),
    }
\end{equation}
where $C^{\left(l\right)}$ are node (concept) features and $W^{\left( l\right)}$ trainable weight matrix of layer $l \in [1, L]$, $\sigma$ is a nonlinear activation function, and $A$ is the adjacency matrix obtained by normalizing graph $G_C$ with the degree matrix $D$. Since $G_C$ is a directed graph, normalization can be formulated as $A\!=\!D^{-1}G_{C}$.

\paragraph{Commonsense Information.}
We use ConceptNet \cite{speer_conceptnet_2017}, a popular knowledge graph that provides information spanning various types of relationships such as physical, spatial, behavioral, \etc To ensure that the ConceptNet information utilized is relevant to themes found in the video data, we consider the set of objects available in pseudo-queries and include the top-$k$ most frequently occurring objects to be the seed concept set \(C\). We extract the  ConceptNet subgraph that includes all edges incident between the concepts in \(C\). 
We filter the edge types based on a pre-determined relation set \(R\), which is compiled to involve relations that are relevant to the nature of the video localization task, \eg, spatial (\textit{AtLocation}, \etc) and temporal (\textit{HasSubevent}, \etc) relations are useful for video understanding, while \textit{RelatedTo} and \textit{Synonym} are fairly generic relations that add little information to the localization task. Table \ref{tab:relations} shows the relations included in \(G_C\).

\paragraph{Cross-Modal Interaction Module.} The commonsense enriched video and query features, \(V_{C}\) and \(Q_{C}\), are fused with a multi-modal cross-attention mechanism. We employ a two-step fusion process. First, Query-guided Video Attention (QVA) is applied to attend over video $V_C$, and Video-guided Query Attention (VQA) attends over query $Q_C$ guided by video $V_C$, resulting in updated features $V_C'$ and $Q_C'$, respectively. Both QVA and VQA utilize Attention Dynamic Filters~\cite{rodriguez_proposal-free_2020} that adaptively modify video features, dynamically adjusting them in response to the query, and vice versa. Next, the attended features are fused using a cross-attention mechanism over $V_C'$ guided by $Q_C'$, resulting in localized video features $V_{C_{\text{loc}}}$.

\paragraph{Temporal Regression Module.}
The final step involves a regression layer that approximates $\hat{V}_{\text{span}}$. We employ attention-guided temporal regression to estimate the span of the target video moment. To find important temporal segments relevant to the query, the fused features $V_{C_{\text{loc}}}$ are temporally attended and span boundaries are localized using a regressor implemented as a Multi-Layer Perceptron (MLP).
\begin{align}
{a}_i = \sigma\left({W}_{1} V_{C_{\text{loc}_i}} + {b}_{{1}}\right) \\
v_{\text{ta}} = \sum_{i=1}^{T} a_i V_{C_{\text{loc}_{i}}} \\
[\hat{t}_s, \hat{t}_e] = {W}_2 {v}_{\text{ta}} + {b}_{2}.
\end{align}
Here, ${W}_{1}$ and ${b}_1$ are the weight matrix and bias vector of the temporal attention MLP, $\sigma$ represents the sigmoid activation function, $V_{C_{\text{loc}_i}}$ stands for the $i$-th localized video feature, ${v}_{\text{ta}}$ represents the temporally attended video feature vector, ${W}_2$ and ${b}_2$ denote the weight matrix and bias vector of the regression MLP, and $[\hat{t}_s, \hat{t}_e]$ correspond to the start and end timestamps of the predicted video span $\hat{V}_{\text{span}}$.

\begin{table}[t!]
    \centering
    \begin{tabular}{ll}
    \toprule
    \textbf{Category} & \textbf{Relations}                                                                                         \\ \toprule
    Spatial           & AtLocation, LocatedNear                                                                                    \\ \midrule
    Temporal          & \begin{tabular}[c]{@{}l@{}}HasSubevent, HasFirstSubevent,\\HasLastSubevent, HasPrerequisite\end{tabular} \\ \midrule
    Functional        & UsedFor                                                                                                    \\ \midrule
    Causal            & Causes                                                                                                     \\ \midrule
    Motivation        & MotivatedByGoal,  ObstructedBy                                                                             \\ \midrule
    Other             & CreatedBy, MadeOf                                                                                          \\ \midrule
    Physical          & \begin{tabular}[c]{@{}l@{}}HasA, HasProperty, Antonym, SimilarTo\end{tabular}
    \\ \bottomrule
    \end{tabular}

    \caption{Relations in the Commonsense Enhancement Module (CEM) grouped by categories.}
    \label{tab:relations}

\end{table}

\subsection{Training and Inference}
The training objective is 
$\mathcal{L}_{loc} = \mathcal{L}_{treg}+\lambda \mathcal{L}_{ta},$ where \(\lambda\) is a balancing hyperparameter, \(\mathcal{L}_{ta}\) is a temporal attention guided loss and \(\mathcal{L}_{treg}\) is the regression loss.  The temporal attention-guided loss is defined as
\begin{equation}
\label{tatt}
\mathcal{L}_{ta} = \frac{\sum^{T}_{i=1}g_{i}\log \left( a_{i}\right)}{\sum^{T}_{i=1}g_{i}},
\end{equation}
where \(a_{i}\) is the attention weight for video frame \(v_{i}\) and \(g_{i}\) is the attention mask for \(v_{i}\), that is assigned to \(1\) if \(v_{i}\) is inside the target video segment, and \(0\) otherwise. 
This objective encourages the model to produce higher attention weights for video segments that are relevant to the query. 
On the other hand, \(\mathcal{L}_{treg}\) dictates the video span boundary regression and is the sum of smooth $\ell_1$ distances between start and end timestamps of the ground truth and predicted spans, \ie,
\begin{equation}
\label{treg}
\mathcal{L}_{treg} = \text{smooth}{\ell_1}(t_{s}, \hat{t}_{s}) + \text{smooth}{\ell_1}(t_{e}, \hat{t}_{e}).
\end{equation}
Here, $t_{s}$ and ${t}_{e}$ represent the ground truth start and end timestamps and $\hat{t}_{s}$ and $\hat{t}_{e}$ the predicted start and end timestamps, respectively.
The integration of a smoothing mechanism enhances training stability and improves the model's ability to handle outliers. Finally, during inference, we employ an off-the-shelf part-of-speech tagger to extract nouns from the text input query and feed them as query input to the trained \modelname video localizer.
\begin{table*}[t!]
\resizebox{\linewidth}{!}{
\begin{tabular}{lcllll||llll}
\toprule
& & \multicolumn{4}{c}{\textbf{Charades-STA}} & \multicolumn{4}{c}{\textbf{ActivityNet-Captions}} \\
\cmidrule(lr){3-6} \cmidrule(lr){7-10}
\textbf{Approach} & \textbf{Supervision} & \textbf{R@0.3} & \textbf{R@0.5} & \textbf{R@0.7} & \textbf{mIoU} & \textbf{R@0.3} & \textbf{R@0.5} & \textbf{R@0.7} & \textbf{mIoU} \\ \midrule
CTRL~\cite{Gao_2017_ICCV} & \multirow{2}{*}{Full} & - & 21.42 & 7.15 & - & 28.70 & 14.00 & - & 20.54 \\
LGI~\cite{mun_local-global_2020} & & 72.96 & 59.46 & 35.48 & 51.38 & 58.52 & 41.51 & 23.07 & 41.13 \\
\midrule
TGA~(Mithun et al. 2019) & \multirow{5}{*}{Weak} & 29.68 & 17.04 & 6.93 & - & - & - & - & - \\
WSTG~\cite{chen_look_2020} & & 39.80 & 27.30 & 12.90 & 27.30 & 44.30 & 23.60 & - & 32.20 \\
SCN~\cite{lin_weakly-supervised_2020} & & 42.96 & 23.58 & 9.97 & - & 47.23 & 29.22 & - & - \\
WS-DEC~\cite{duan2018weakly} & & - & - & - & - & 41.98 & 23.34 & - & 28.23 \\
WSLLN~\cite{gao2019wslln} & & - & - & - & - & 42.80 & 22.70 & - & 32.20 \\
\midrule
PSVL$^{\dagger}$~\cite{nam_zero-shot_2021} & \multirow{4}{*}{None} & 46.63 & 30.84 & 13.57 & 31.09 & 43.03 & 25.14 & 10.96 & 30.77 \\
LFVL$^{\dagger}$~\cite{kim2023language} & & \underline{49.50} & \underline{34.39} & \underline{16.95} & \textbf{33.19} & 43.34 & 25.17 & \textbf{13.10} & \textbf{31.67} \\
\textbf{\modelname} & & {49.21} & \textbf{34.60} & \textbf{17.93} & {{32.73}} & {\textbf{46.05}} & {\underline{28.19}} & \underline{12.84} & \underline{31.11} \\
\textbf{\modelname$_{250}$} & & {\textbf{50.98}} & {{33.18}} & {{16.48}} & \underline{33.06} & \underline{45.43} & \textbf{28.27} & 12.81 &{30.88} \\
\bottomrule
\end{tabular}
}
\caption{Localization accuracy compared to zero-shot, weakly, and fully supervised baselines. $^{\dagger}$ indicates reproduction with official checkpoints and/or implementations. Best-performing method is highlighted in bold and second-best is underlined.}
\label{tab:resultsCompare}
\end{table*}
\section{Experiments}

\paragraph{Experimental Setup.}
Consistent with prior zero-shot NLVL research, we evaluate on Charades-STA \cite{Gao_2017_ICCV} and ActivityNet-Captions \cite{heilbron2015activitynet,krishna2017dense}. Note that we only utilize the video components of the dataset during training. Query and video span annotations are only used for evaluation purposes. We compare \modelname against several zero-shot \cite{nam_zero-shot_2021,kim2023language}, weakly supervised \cite{mithun_weakly_2019,chen_look_2020,lin_weakly-supervised_2020,duan2018weakly,gao2019wslln} and fully supervised~\cite{Gao_2017_ICCV,mun_local-global_2020} baselines. We evaluate performance with the mean temporal Intersection over Union (\textit{mIoU}) for the predicted video moment spans and recall at specific threshold values (\textit{R@k}), which is defined as the percentage of video span predictions with IoU value of at least \(k\), where \(k= \left\{ 0.3,0.5,0.7\right\} \), following prior works. 

\subsection{Experimental Results}
Table \ref{tab:resultsCompare} illustrates a comparative analysis of \modelname against baselines. We compare \modelname using two  \(G_{C}\) versions with varying the number of concepts in the commonsense module, \ie \(n_C \in \{300, 250\}\). We represent these configurations as \modelname and \modelname$_{250}$, respectively. 
\modelname outperforms the fully supervised CTRL baseline and all of the weakly supervised baselines by significant margins. In addition, \modelname surpasses the PSVL zero-shot baseline across various configurations, with a particularly strong performance in the higher recall regime (\(R@0.7\)).
For instance, for the Charades-STA dataset, \modelname consistently outperforms PSVL, yielding gains of up to $32.13\%$ in higher recall scenarios. Similarly, on the same dataset, \modelname achieves recall enhancements of up to $5.78\%$ over LFVL for \(R@0.7\). In the context of the ActivityNet-Captions dataset, \modelname also outperforms PSVL across all metrics, showcasing performance improvements ranging from $7.02\%$ to $17.15\%$.
Notably, \modelname substantially outperforms LFVL on ActivityNet-Captions in terms of \(R@0.3\) and \(R@0.5\) (up to $12\%$ for \(R@0.5\)), while maintaining comparable results in terms of \(R@0.7\) and \(mIoU\). Considering that ActivityNet-Captions represents a challenging benchmark encompassing diverse video themes, our findings highlight that leveraging commonsense information effectively helps integrate diverse visual-linguistic themes, outperforming methods that rely on pre-trained large-scale vision-language models.

Furthermore, since \(mIoU\) for all models is close to 30\%, an increase in  \(mIoU\) corresponds to a proportional increase in model predictions with recall above 0.3 (\(R@0.3\)).
The performance of \modelname with lower \vs higher $G_{C}$ sizes highlights a pattern of exclusivity between overall localization performance (\ie, \(mIoU\)) and precision of accuracy (higher recall regimes, \eg, \(R@0.7\)). This dichotomy between better recall at higher regimes and increased average localization highlights the trade-off between being able to generalize to a diverse set of videos and accurately localizing the moment in a specific video. Higher concept set sizes may provide wider levels of information to accommodate different types of videos better but may impede the model's capability to ground the exact video moment accurately. 

\begin{table}[t!]
\centering
\begin{tabular}{lrrrr}
\toprule
\textbf{Relations} & \multicolumn{1}{l}{\textbf{R@0.3}} & \multicolumn{1}{l}{\textbf{R@0.5}} & \multicolumn{1}{l}{\textbf{R@0.7}} & \multicolumn{1}{l}{\textbf{mIoU}} \\ \midrule
\textbf{S}         & 39.63                              & 26.53                              & 11.81                              & 26.03                             \\
\textbf{T}         & 44.98                              & 28.08                              & 13.93                              & 29.63                             \\
\textbf{ST}        & 49.84                              & 30.35                              & 15.16                              & 32.38                             \\
\textbf{F}         & 49.21                              & 34.60                               & 17.93                              & 32.73                             \\
\textbf{F-ST}    &  47.98	& 29.26&	14.29&	30.77\\
\textbf{All}       & 49.42                              & 34.03                              & 17.99                              & 32.85   \\
\bottomrule
\end{tabular}

\caption{\modelname with Spatial (\textbf{S}), Temporal (\textbf{T}), and Spatial and Temporal (\textbf{ST}) relations, the customized set of Filtered relations (\textbf{F}) mentioned in Table \ref{tab:relations}, \textbf{F} without the spatial/temporal relations (\textbf{F-ST}) and \textbf{All} relation types.}

\label{tab:ablationRelation}
\end{table}

\subsection{Ablation Studies}
\label{sec:ablation}
Comprehensive ablation studies, found in the appendix, provide further insights into the importance of commonsense in zero-shot NLVL. Specifically, we evaluate (1) the influence of various relation types, (2) the relative significance of commonsense in augmenting video or query features, (3) the potential usefulness of auxiliary commonsense information, and (4) the best approach for injecting commonsense. 

\begin{figure*}[t!]
    \centering
    \includegraphics[width=.95\linewidth]{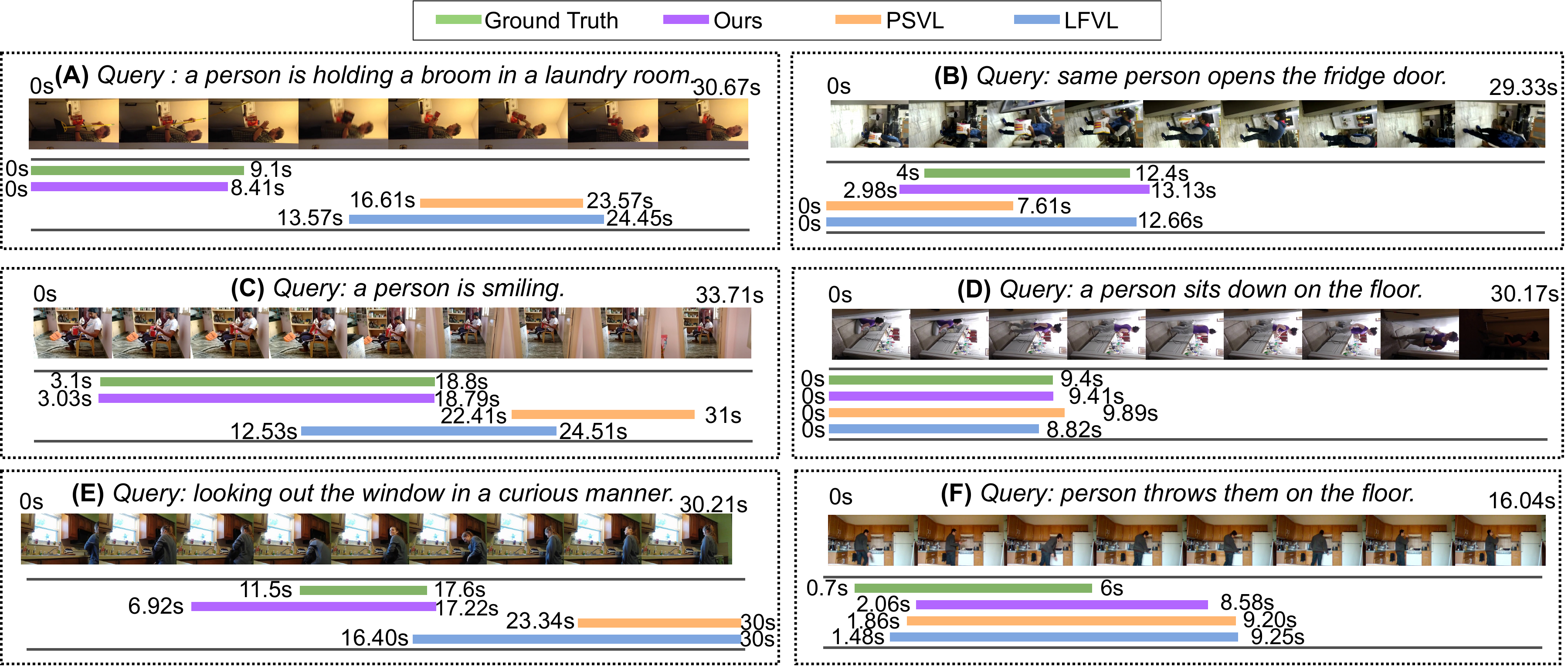}
  \caption{Qualitative inference results on examples from Charades-STA test data. Video span timestamps predicted by \modelname (\textcolor[HTML]{9900FF}{{purple}} lines), PSVL (\textcolor[HTML]{FF9900}{{orange}} lines), and LFVL (\textcolor[HTML]{7EA6E0}{{blue}} lines), juxtaposed with ground truth timestamps (\textcolor[HTML]{82B366}{{green}} lines).
  }
  \label{fig:qual}
\end{figure*}

\subsubsection{Which Relation Types Are Most Important?}
We analyze the contribution and relative importance of relation types used in building \(G_{C}\). Given that video localization requires spatiotemporal understanding, we hypothesize that relations falling in spatial/temporal categories are essential. Accordingly, we examine the performance of \modelname with 1) only spatial relations (\textbf{S}), 2) only temporal relations (\textbf{T}), and 3) both spatial and temporal relations (\textbf{ST}). We also evaluate \modelname with 4) a bigger subset of relations (as given in Table \ref{tab:relations}) to accommodate domain invariance (\textbf{F}), 5) the relation set mentioned in the previous configuration excluding spatial and temporal relation types (\textbf{F-ST}), and 6) all relations in ConceptNet (\textbf{All}). 

Table \ref{tab:ablationRelation} enumerates the results across all \modelname configurations. 
The performance drops significantly across all metrics with \textbf{S}, where only spatial relations are considered. The temporal relation set \textbf{T} performs much better than \textbf{S}, highlighting the higher importance of temporal commonsense than spatial commonsense for precise localization. Despite the poor recall performance on \textbf{S}, spatial relations are valuable to the localization process, which is supported by a further improved performance with \textbf{ST}, which considers both spatial and temporal relations. The considerable drop in performance of \textbf{F-ST}, a filtered set that excludes spatial and temporal relation types, as compared to \textbf{F} and \textbf{ST}, further emphasizes the importance of spatial and temporal commonsense information for accurate localization. 
Finally, performance is considerably high with all relations included (\textbf{All}), but not significantly better than previous configurations that use a far smaller $G_{C}$ and are hence more resource-efficient. Overall, \textbf{F} seems to provide a good balance between recall at various levels and mean localization accuracy. 

\section{Qualitative Results}
We present qualitative results w.r.t. \modelname's localization performance along with the PSVL and LFVL baselines. In Figure~\ref{fig:qual}, we showcase a few examples from the Charades-STA test split, accompanied by ground-truth video and query annotations and the localization results of the three models. Examples (A)-(C) show how \modelname accurately localizes the video moment, while PSVL and LFVL perform poorly. Upon inspection, PSVL and LFVL localize succeeding but semantically different events from the target segment in (A) and (C). 
On the other hand, in (B), PSVL and LFVL are seen to localize temporally preceding events along with the target event jointly. 
This example is challenging since the query includes ``same person", which requires the ability to contextualize and distinguish preceding events from the target event. The inability of PSVL and LFVL baselines to localize the target segment alone (\ie, isolating it from the preceding contextual segments), and conversely the accurate localization performance of \modelname, shows that \modelname can effectively contextualize and distinguish ``same person" as a co-referenced entity by leveraging commonsense. Our results corroborate previous works that demonstrate the usefulness of commonsense in co-reference resolution~\cite{coref_commonsense_ieee, coref_commonsense_acl_ravi-etal-2023-happens}. 

Example (D) highlights a case where all three models can localize accurately, while (E) and (F) show examples where none of the three models performs exceptionally well. However, it is important to note that \modelname localizes closest to the ground truth -- it captures the ground truth event, but also jointly localizes the preceding event, which is semantically similar to the target event, \ie, the person is looking outside the window. In contrast, PSVL and LFVL localize video segments that showcase events that are very distinct from the target query, \ie, the person in the frame is looking away from the window (\eg, LFVL localizes an event where the person is looking at the camera, whereas PSVL localizes the succeeding event where the person is looking down). Finally, all three models perform similarly in example (F). A deeper analysis reveals that each model localizes neighboring events in addition to the target event. 
\section{Conclusion}
In this paper, we integrate commonsense in zero-shot natural language video localization to reduce noise in pseudo-queries and enhance cross-modal grounding between video and query modalities. Experimental results demonstrate the impact of commonsense relational information in enriching video and query representations, resulting in improved recall and localization performance within the zero-shot setting.

\section{Acknowledgements}\label{acknowledgement}
This material is based upon work supported in part by the U.S. DARPA KMASS Program \#HR001121S0034, the Amazon – Virginia Tech Initiative for Efficient and Robust Machine Learning, and the Amazon Alexa Prize TaskBot Challenge 2.  
The U.S. Government is authorized to reproduce and distribute reprints for Governmental purposes notwithstanding any copyright notation thereon.
The views and conclusions contained herein are those of the authors and should not be interpreted as necessarily representing the official policies or endorsements - either expressed or implied - of Amazon, DARPA, or the U.S. Government.

{\bibliography{main}}

\begin{thebibliography}{62}
\providecommand{\natexlab}[1]{#1}

\bibitem[{Cao et~al.(2022)Cao, Wang, Zhang, and Ma}]{cao_visual_2022}
Cao, S.; Wang, B.; Zhang, W.; and Ma, L. 2022.
\newblock Visual {Consensus} {Modeling} for {Video}-{Text} {Retrieval}.
\newblock In \emph{AAAI}, 167--175.

\bibitem[{Chen and Jiang(2020)}]{chen_hierarchical_2020}
Chen, S.; and Jiang, Y.-G. 2020.
\newblock Hierarchical {Visual}-{Textual} {Graph} for {Temporal} {Activity} {Localization} via {Language}.
\newblock In \emph{ECCV}, 601--618.

\bibitem[{Chen et~al.(2020)Chen, Ma, Luo, Tang, and Wong}]{chen_look_2020}
Chen, Z.; Ma, L.; Luo, W.; Tang, P.; and Wong, K.-Y.~K. 2020.
\newblock Look {Closer} to {Ground} {Better}: {Weakly}-{Supervised} {Temporal} {Grounding} of {Sentence} in {Video}.
\newblock \emph{ArXiv:2001.09308}.

\bibitem[{Ding et~al.(2021)Ding, Chen, Du, Luo, Tenenbaum, and Gan}]{ding_dynamic_2021}
Ding, M.; Chen, Z.; Du, T.; Luo, P.; Tenenbaum, J.; and Gan, C. 2021.
\newblock Dynamic {Visual} {Reasoning} by {Learning} {Differentiable} {Physics} {Models} from {Video} and {Language}.
\newblock In \emph{NeurIPS}, 887--899.

\bibitem[{Duan et~al.(2018)Duan, Huang, Gan, Wang, Zhu, and Huang}]{duan2018weakly}
Duan, X.; Huang, W.; Gan, C.; Wang, J.; Zhu, W.; and Huang, J. 2018.
\newblock {Weakly Supervised Dense Event Captioning in Videos}.
\newblock In \emph{NeurIPS}, 3062--3072.

\bibitem[{Fang et~al.(2020)Fang, Gokhale, Banerjee, Baral, and Yang}]{fang_video2commonsense_2020}
Fang, Z.; Gokhale, T.; Banerjee, P.; Baral, C.; and Yang, Y. 2020.
\newblock {Video2Commonsense}: {Generating} {Commonsense} {Descriptions} to {Enrich} {Video} {Captioning}.
\newblock In \emph{EMNLP}, 840--860.

\bibitem[{Gao et~al.(2017)Gao, Sun, Yang, and Nevatia}]{Gao_2017_ICCV}
Gao, J.; Sun, C.; Yang, Z.; and Nevatia, R. 2017.
\newblock {TALL: Temporal Activity Localization via Language Query}.
\newblock In \emph{ICCV}.

\bibitem[{Gao et~al.(2021)Gao, Sun, Xu, Zhou, and Ghanem}]{gao_relation-aware_2021}
Gao, J.; Sun, X.; Xu, M.; Zhou, X.; and Ghanem, B. 2021.
\newblock Relation-aware {Video} {Reading} {Comprehension} for {Temporal} {Language} {Grounding}.
\newblock In \emph{EMNLP}, 3978--3988.

\bibitem[{Gao and Xu(2021)}]{gao_fast_2021}
Gao, J.; and Xu, C. 2021.
\newblock Fast Video Moment Retrieval.
\newblock In \emph{ICCV}, 1503--1512.

\bibitem[{Gao et~al.(2019)Gao, Davis, Socher, and Xiong}]{gao2019wslln}
Gao, M.; Davis, L.; Socher, R.; and Xiong, C. 2019.
\newblock {{WSLLN}:Weakly Supervised Natural Language Localization Networks}.
\newblock In \emph{EMNLP-IJCNLP}, 1481--1487.

\bibitem[{Gao et~al.(2022)Gao, Luo, Chen, and Zhou}]{gao_end--end_2022}
Gao, Y.; Luo, Z.; Chen, S.; and Zhou, W. 2022.
\newblock End-to-end {Multi}-task {Learning} {Framework} for {Spatio}-{Temporal} {Grounding} in {Video} {Corpus}.
\newblock In \emph{CIKM}, 3958--3962.

\bibitem[{Goyal et~al.(2017)Goyal, Ebrahimi~Kahou, Michalski, Materzynska, Westphal, Kim, Haenel, Fruend, Yianilos, Mueller-Freitag, Hoppe, Thurau, Bax, and Memisevic}]{goyal_something_2017}
Goyal, R.; Ebrahimi~Kahou, S.; Michalski, V.; Materzynska, J.; Westphal, S.; Kim, H.; Haenel, V.; Fruend, I.; Yianilos, P.; Mueller-Freitag, M.; Hoppe, F.; Thurau, C.; Bax, I.; and Memisevic, R. 2017.
\newblock The "{Something} {Something}" {Video} {Database} for {Learning} and {Evaluating} {Visual} {Common} {Sense}.
\newblock In \emph{ICCV}, 5842--5850.

\bibitem[{Hammond, Vandergheynst, and Gribonval(2011)}]{hammond_wavelets_2011}
Hammond, D.~K.; Vandergheynst, P.; and Gribonval, R. 2011.
\newblock Wavelets on graphs via spectral graph theory.
\newblock \emph{Applied and Computational Harmonic Analysis}, 30: 129--150.

\bibitem[{He et~al.(2022)He, Mao, Zhou, Li, Gong, Li, and Wu}]{coref_commonsense_ieee}
He, K.; Mao, B.; Zhou, X.; Li, Y.; Gong, T.; Li, C.; and Wu, J. 2022.
\newblock {Knowledge Enhanced Coreference Resolution via Gated Attention}.
\newblock In \emph{IEEE BIBM}, 2287--2293.

\bibitem[{Heilbron et~al.(2015)Heilbron, Escorcia, Ghanem, and Niebles}]{heilbron2015activitynet}
Heilbron, F.~C.; Escorcia, V.; Ghanem, B.; and Niebles, J.~C. 2015.
\newblock {ActivityNet: A Large-scale Video Benchmark for Human Activity Understanding}.
\newblock In \emph{CVPR}.

\bibitem[{Huang et~al.(2021)Huang, Liu, Gong, and Jin}]{huang_cross-sentence_2021}
Huang, J.; Liu, Y.; Gong, S.; and Jin, H. 2021.
\newblock {Cross-Sentence Temporal and Semantic Relations in Video Activity Localisation}.
\newblock In \emph{ICCV)}, 7199--7208.

\bibitem[{Jiang et~al.(2022)Jiang, Lin, Han, Song, and Huang}]{jiang_pseudo-q_2022}
Jiang, H.; Lin, Y.; Han, D.; Song, S.; and Huang, G. 2022.
\newblock Pseudo-{Q}: {Generating} {Pseudo} {Language} {Queries} for {Visual} {Grounding}.
\newblock In \emph{CVPR}, 15513--15523.

\bibitem[{Kim et~al.(2023)Kim, Park, Lee, Park, and Sohn}]{kim2023language}
Kim, D.; Park, J.; Lee, J.; Park, S.; and Sohn, K. 2023.
\newblock Language-free Training for Zero-shot Video Grounding.
\newblock In \emph{WACV}, 2539--2548.

\bibitem[{Krishna et~al.(2017{\natexlab{a}})Krishna, Hata, Ren, Fei-Fei, and Niebles}]{krishna2017dense}
Krishna, R.; Hata, K.; Ren, F.; Fei-Fei, L.; and Niebles, J.~C. 2017{\natexlab{a}}.
\newblock {Dense-Captioning Events in Videos}.
\newblock In \emph{ICCV}.

\bibitem[{Krishna et~al.(2017{\natexlab{b}})Krishna, Zhu, Groth, Johnson, Hata, Kravitz, Chen, Kalantidis, Li, Shamma, Bernstein, and Fei-Fei}]{krishna_visual_2017}
Krishna, R.; Zhu, Y.; Groth, O.; Johnson, J.; Hata, K.; Kravitz, J.; Chen, S.; Kalantidis, Y.; Li, L.-J.; Shamma, D.~A.; Bernstein, M.~S.; and Fei-Fei, L. 2017{\natexlab{b}}.
\newblock Visual {Genome}: {Connecting} {Language} and {Vision} {Using} {Crowdsourced} {Dense} {Image} {Annotations}.
\newblock \emph{International Journal of Computer Vision}, 123: 32--73.

\bibitem[{Lei, Berg, and Bansal(2021)}]{detr}
Lei, J.; Berg, T.~L.; and Bansal, M. 2021.
\newblock Detecting Moments and Highlights in Videos via Natural Language Queries.
\newblock In Ranzato, M.; Beygelzimer, A.; Dauphin, Y.; Liang, P.; and Vaughan, J.~W., eds., \emph{NeurIPS}, 11846--11858.

\bibitem[{Lei et~al.(2020{\natexlab{a}})Lei, Yu, Berg, and Bansal}]{lei_tvqa_2020}
Lei, J.; Yu, L.; Berg, T.; and Bansal, M. 2020{\natexlab{a}}.
\newblock {TVQA}+: {Spatio}-{Temporal} {Grounding} for {Video} {Question} {Answering}.
\newblock In \emph{ACL}, 8211--8225.

\bibitem[{Lei et~al.(2020{\natexlab{b}})Lei, Yu, Berg, and Bansal}]{lei_what_2020}
Lei, J.; Yu, L.; Berg, T.; and Bansal, M. 2020{\natexlab{b}}.
\newblock What is {More} {Likely} to {Happen} {Next}? {Video}-and-{Language} {Future} {Event} {Prediction}.
\newblock In \emph{EMNLP}, 8769--8784.

\bibitem[{Li, Niu, and Zhang(2022)}]{li_representation_2022}
Li, J.; Niu, L.; and Zhang, L. 2022.
\newblock From {Representation} to {Reasoning}: {Towards} {Both} {Evidence} and {Commonsense} {Reasoning} for {Video} {Question}-{Answering}.
\newblock In \emph{CVPR}, 21273--21282.

\bibitem[{Li, Guo, and Wang(2021)}]{li_proposal-free_2021}
Li, K.; Guo, D.; and Wang, M. 2021.
\newblock {Proposal-free Video Grounding with Contextual Pyramid Network}.
\newblock In \emph{AAAI}, 1902--1910.

\bibitem[{Li et~al.(2023)Li, Xie, Xie, Zhao, Zhang, Zheng, Zhao, and Zhang}]{li2023momentdiff}
Li, P.; Xie, C.-W.; Xie, H.; Zhao, L.; Zhang, L.; Zheng, Y.; Zhao, D.; and Zhang, Y. 2023.
\newblock MomentDiff: Generative Video Moment Retrieval from Random to Real.
\newblock In \emph{NeurIPS}.

\bibitem[{Lin et~al.(2020)Lin, Zhao, Zhang, Wang, and Liu}]{lin_weakly-supervised_2020}
Lin, Z.; Zhao, Z.; Zhang, Z.; Wang, Q.; and Liu, H. 2020.
\newblock Weakly-{Supervised} {Video} {Moment} {Retrieval} via {Semantic} {Completion} {Network}.
\newblock In \emph{{AAAI}}.

\bibitem[{Liu et~al.(2021)Liu, Qu, Dong, Zhou, Cheng, Wei, Xu, and Xie}]{liu_context-aware_2021}
Liu, D.; Qu, X.; Dong, J.; Zhou, P.; Cheng, Y.; Wei, W.; Xu, Z.; and Xie, Y. 2021.
\newblock Context-{Aware} {Biaffine} {Localizing} {Network} for {Temporal} {Sentence} {Grounding}.
\newblock In \emph{CVPR}, 11235--11244.

\bibitem[{Liu et~al.(2022{\natexlab{a}})Liu, Qu, Wang, Di, Zou, Cheng, Xu, and Zhou}]{liu_unsupervised_2022}
Liu, D.; Qu, X.; Wang, Y.; Di, X.; Zou, K.; Cheng, Y.; Xu, Z.; and Zhou, P. 2022{\natexlab{a}}.
\newblock Unsupervised {Temporal} {Video} {Grounding} with {Deep} {Semantic} {Clustering}.
\newblock In \emph{AAAI}, 1683--1691.

\bibitem[{Liu et~al.(2022{\natexlab{b}})Liu, Qu, Zhou, and Liu}]{liu_exploring_2022}
Liu, D.; Qu, X.; Zhou, P.; and Liu, Y. 2022{\natexlab{b}}.
\newblock Exploring {Motion} and {Appearance} {Information} for {Temporal} {Sentence} {Grounding}.
\newblock In \emph{AAAI}, 1674--1682.

\bibitem[{Liu et~al.(2020)Liu, Chen, Cheng, Gan, Yu, Yang, and Liu}]{liu_violin_2020}
Liu, J.; Chen, W.; Cheng, Y.; Gan, Z.; Yu, L.; Yang, Y.; and Liu, J. 2020.
\newblock Violin: {A} {Large}-{Scale} {Dataset} for {Video}-and-{Language} {Inference}.
\newblock In \emph{CVPR}, 10900--10910.

\bibitem[{Ma et~al.(2020)Ma, Yoon, Kim, Lee, Kang, and Yoo}]{ma_vlanet_2020}
Ma, M.; Yoon, S.; Kim, J.; Lee, Y.; Kang, S.; and Yoo, C.~D. 2020.
\newblock {VLANet}: {Video}-{Language} {Alignment} {Network} for {Weakly}-{Supervised} {Video} {Moment} {Retrieval}.
\newblock In \emph{ECCV}, 156--171.

\bibitem[{Maharana and Bansal(2021)}]{maharana_integrating_2021}
Maharana, A.; and Bansal, M. 2021.
\newblock Integrating {Visuospatial}, {Linguistic}, and {Commonsense} {Structure} into {Story} {Visualization}.
\newblock In \emph{EMNLP}, 6772--6786.

\bibitem[{Mithun, Paul, and Roy-Chowdhury(2019)}]{mithun_weakly_2019}
Mithun, N.~C.; Paul, S.; and Roy-Chowdhury, A.~K. 2019.
\newblock {Weakly Supervised Video Moment Retrieval from Text Queries}.
\newblock In \emph{CVPR}, 11592--11601.

\bibitem[{Mun, Cho, and Han(2020)}]{mun_local-global_2020}
Mun, J.; Cho, M.; and Han, B. 2020.
\newblock Local-{Global} {Video}-{Text} {Interactions} for {Temporal} {Grounding}.
\newblock In \emph{CVPR}.

\bibitem[{Nam et~al.(2021)Nam, Ahn, Kang, Ha, and Choi}]{nam_zero-shot_2021}
Nam, J.; Ahn, D.; Kang, D.; Ha, S.~J.; and Choi, J. 2021.
\newblock Zero-{Shot} {Natural} {Language} {Video} {Localization}.
\newblock In \emph{ICCV}, 1470--1479.

\bibitem[{Park et~al.(2022)Park, Shen, Farhadi, Darrell, Choi, and Rohrbach}]{park_exposing_2022}
Park, J.~S.; Shen, S.; Farhadi, A.; Darrell, T.; Choi, Y.; and Rohrbach, A. 2022.
\newblock Exposing the {Limits} of {Video}-{Text} {Models} through {Contrast} {Sets}.
\newblock In \emph{NAACL-HTL}, 3574--3586.

\bibitem[{Peng et~al.(2021)Peng, Huang, Xu, Li, Liu, Rahmani, Ke, Guo, Wu, Li, Ye, Wang, Zhang, Liu, He, Zhang, Liu, and Lin}]{peng_multi-modal_2021}
Peng, H.; Huang, H.; Xu, L.; Li, T.; Liu, J.; Rahmani, H.; Ke, Q.; Guo, Z.; Wu, C.; Li, R.; Ye, M.; Wang, J.; Zhang, J.; Liu, Y.; He, T.; Zhang, F.; Liu, X.; and Lin, T. 2021.
\newblock {The Multi-Modal Video Reasoning and Analyzing Competition}.
\newblock In \emph{ICCV Workshops (ICCVW)}, 806--813.

\bibitem[{Qian et~al.(2023)Qian, Cui, Chen, Peng, Guo, and Jiang}]{qian_locate_2022}
Qian, T.; Cui, R.; Chen, J.; Peng, P.; Guo, X.; and Jiang, Y.-G. 2023.
\newblock Locate before {Answering}: {Answer} {Guided} {Question} {Localization} for {Video} {Question} {Answering}.
\newblock \emph{IEEE Transactions on Multimedia}.

\bibitem[{Radford et~al.(2021)Radford, Kim, Hallacy, Ramesh, Goh, Agarwal, Sastry, Askell, Mishkin, Clark et~al.}]{radford2021learning}
Radford, A.; Kim, J.~W.; Hallacy, C.; Ramesh, A.; Goh, G.; Agarwal, S.; Sastry, G.; Askell, A.; Mishkin, P.; Clark, J.; et~al. 2021.
\newblock {Learning Transferable Visual Models from Natural Language Supervision}.
\newblock In \emph{ICML}, 8748--8763.

\bibitem[{Ravi et~al.(2023)Ravi, Tanner, Ng, and Shwartz}]{coref_commonsense_acl_ravi-etal-2023-happens}
Ravi, S.; Tanner, C.; Ng, R.; and Shwartz, V. 2023.
\newblock {What happens before and after: Multi-Event Commonsense in Event Coreference Resolution}.
\newblock In \emph{EACL}, 1708--1724.

\bibitem[{Rodriguez et~al.(2020)Rodriguez, Marrese-Taylor, Saleh, LI, and Gould}]{rodriguez_proposal-free_2020}
Rodriguez, C.; Marrese-Taylor, E.; Saleh, F.~S.; LI, H.; and Gould, S. 2020.
\newblock Proposal-free {Temporal} {Moment} {Localization} of a {Natural}-{Language} {Query} in {Video} using {Guided} {Attention}.
\newblock In \emph{WACV}.

\bibitem[{Rodriguez-Opazo et~al.(2021)Rodriguez-Opazo, Marrese-Taylor, Fernando, Li, and Gould}]{rodriguez-opazo_dori_2021}
Rodriguez-Opazo, C.; Marrese-Taylor, E.; Fernando, B.; Li, H.; and Gould, S. 2021.
\newblock {DORi: Discovering Object Relationships for Moment Localization of a Natural Language Query in a Video}.
\newblock In \emph{WACV}, 1079--1088.

\bibitem[{Shin et~al.(2021)Shin, Kim, Choi, Heo, Kim, Lee, Zhang, and Ryu}]{shin_cogme_2021}
Shin, M.; Kim, J.; Choi, S.; Heo, Y.; Kim, D.; Lee, M.~S.; Zhang, B.; and Ryu, J. 2021.
\newblock {CogME: {A} Novel Evaluation Metric for Video Understanding Intelligence}.
\newblock \emph{ArXiv:2107.09847}.

\bibitem[{Soldan et~al.(2021)Soldan, Xu, Qu, Tegner, and Ghanem}]{soldan_vlg-net_2021}
Soldan, M.; Xu, M.; Qu, S.; Tegner, J.; and Ghanem, B. 2021.
\newblock {VLG}-{Net}: {Video}-{Language} {Graph} {Matching} {Network} for {Video} {Grounding}.
\newblock In \emph{ICCV}, 3224--3234.

\bibitem[{Speer, Chin, and Havasi(2017)}]{speer_conceptnet_2017}
Speer, R.; Chin, J.; and Havasi, C. 2017.
\newblock {ConceptNet} 5.5: {An} {Open} {Multilingual} {Graph} of {General} {Knowledge}.
\newblock In \emph{AAAI}, 4444--4451.

\bibitem[{Tan et~al.(2021)Tan, Xu, Saenko, and Plummer}]{tan_logan_2021}
Tan, R.; Xu, H.; Saenko, K.; and Plummer, B.~A. 2021.
\newblock {LoGAN}: {Latent} {Graph} {Co}-{Attention} {Network} for {Weakly}-{Supervised} {Video} {Moment} {Retrieval}.
\newblock In \emph{WACV}, 2083--2092.

\bibitem[{Tran et~al.(2015)Tran, Bourdev, Fergus, Torresani, and Paluri}]{c3d}
Tran, D.; Bourdev, L.; Fergus, R.; Torresani, L.; and Paluri, M. 2015.
\newblock {Learning Spatiotemporal Features with 3D Convolutional Networks}.
\newblock In \emph{ICCV}, 4489–4497.

\bibitem[{Vaswani et~al.(2017)Vaswani, Shazeer, Parmar, Uszkoreit, Jones, Gomez, Kaiser, and Polosukhin}]{vaswani_attention_2017}
Vaswani, A.; Shazeer, N.; Parmar, N.; Uszkoreit, J.; Jones, L.; Gomez, A.~N.; Kaiser, {\L}.; and Polosukhin, I. 2017.
\newblock Attention is {All} you {Need}.
\newblock In \emph{NeurIPS}.

\bibitem[{Wang, Ma, and Jiang(2020)}]{wang_temporally_2020}
Wang, J.; Ma, L.; and Jiang, W. 2020.
\newblock Temporally {Grounding} {Language} {Queries} in {Videos} by {Contextual} {Boundary}-{Aware} {Prediction}.
\newblock In \emph{AAAI}, 12168--12175.

\bibitem[{Wang, Zhou, and Li(2021)}]{wang_fine-grained_2021}
Wang, Y.; Zhou, W.; and Li, H. 2021.
\newblock {Fine-grained Semantic Alignment Network for Weakly Supervised Temporal Language Grounding}.
\newblock In \emph{Findings of EMNLP}, 89--99.

\bibitem[{Wu et~al.(2022)Wu, Gao, Huang, and Xu}]{wu_learning_2022}
Wu, Z.; Gao, J.; Huang, S.; and Xu, C. 2022.
\newblock Learning {Commonsense}-aware {Moment}-{Text} {Alignment} for {Fast} {Video} {Temporal} {Grounding}.
\newblock \emph{ArXiv:2204.01450}.

\bibitem[{Xiao et~al.(2021)Xiao, Chen, Zhang, Ji, Shao, Ye, and Xiao}]{xiao_boundary_2021}
Xiao, S.; Chen, L.; Zhang, S.; Ji, W.; Shao, J.; Ye, L.; and Xiao, J. 2021.
\newblock {Boundary Proposal Network for Two-Stage Natural Language Video Localization}.
\newblock In \emph{AAAI}.

\bibitem[{Yang et~al.(2021)Yang, Feng, Ji, Wang, and Chua}]{yang_deconfounded_2021}
Yang, X.; Feng, F.; Ji, W.; Wang, M.; and Chua, T.-S. 2021.
\newblock {Deconfounded Video Moment Retrieval with Causal Intervention}.
\newblock In \emph{ACM SIGIR}, 1–10.

\bibitem[{Yu et~al.(2021)Yu, Liang, Ji, Li, Fang, Xiao, and Duan}]{yu_hybrid_2021}
Yu, W.; Liang, J.; Ji, L.; Li, L.; Fang, Y.; Xiao, N.; and Duan, N. 2021.
\newblock Hybrid {Reasoning} {Network} for {Video}-based {Commonsense} {Captioning}.
\newblock In \emph{ACM MM}, 5213--5221.

\bibitem[{Yu et~al.(2020)Yu, Song, Yu, Wang, and Huang}]{yu_intra-_2020}
Yu, Z.; Song, Y.; Yu, J.; Wang, M.; and Huang, Q. 2020.
\newblock Intra- and {Inter}-modal {Multilinear} {Pooling} with {Multitask} {Learning} for {Video} {Grounding}.
\newblock \emph{Neural Processing Letters}, 52(3): 1863--1879.

\bibitem[{Zeng et~al.(2020)Zeng, Xu, Huang, Chen, Tan, and Gan}]{zeng_dense_2020}
Zeng, R.; Xu, H.; Huang, W.; Chen, P.; Tan, M.; and Gan, C. 2020.
\newblock Dense {Regression} {Network} for {Video} {Grounding}.
\newblock In \emph{CVPR}.

\bibitem[{Zeng et~al.(2021)Zeng, Cao, Wei, Liu, Zhao, and Qin}]{zeng_multi-modal_2021}
Zeng, Y.; Cao, D.; Wei, X.; Liu, M.; Zhao, Z.; and Qin, Z. 2021.
\newblock Multi-{Modal} {Relational} {Graph} for {Cross}-{Modal} {Video} {Moment} {Retrieval}.
\newblock In \emph{CVPR}, 2215--2224.

\bibitem[{Zhang et~al.(2021)Zhang, Sun, Jing, Nan, Zhen, Zhou, and Goh}]{zhang_video_2021}
Zhang, H.; Sun, A.; Jing, W.; Nan, G.; Zhen, L.; Zhou, J.~T.; and Goh, R. S.~M. 2021.
\newblock Video {Corpus} {Moment} {Retrieval} with {Contrastive} {Learning}.
\newblock In \emph{ACM SIGIR}, 685--695.

\bibitem[{Zhang and Radke(2022)}]{zhang_natural_2022}
Zhang, L.; and Radke, R.~J. 2022.
\newblock Natural Language Video Moment Localization Through Query-Controlled Temporal Convolution.
\newblock In \emph{WACV}, 2524--2532.

\bibitem[{Zhang et~al.(2020)Zhang, Zhao, Lin, zhu, and He}]{zhang_counterfactual_2020}
Zhang, Z.; Zhao, Z.; Lin, Z.; zhu, j.; and He, X. 2020.
\newblock {Counterfactual Contrastive Learning for Weakly-Supervised Vision-Language Grounding}.
\newblock In \emph{NeurIPS}, 18123--18134.

\bibitem[{Zhao et~al.(2021)Zhao, Zhao, Zhang, and Lin}]{zhao_cascaded_2021}
Zhao, Y.; Zhao, Z.; Zhang, Z.; and Lin, Z. 2021.
\newblock Cascaded Prediction Network via Segment Tree for Temporal Video Grounding.
\newblock In \emph{CVPR}, 4197--4206.

\end{thebibliography}
\clearpage
\newpage
\appendix
\section{Implementation Details}
We employ pre-trained I3D~\cite{qian_locate_2022} and C3D~\cite{c3d} models to extract video frame features for Charades-STA and ActivityNet-Captions, respectively. We uniformly sample $T\!=
\!128$ features per video to ensure a fixed length.
During the pseudo-query generation phase, we employ a Faster R-CNN object detector that is trained on objects enumerated in VisualGenome~\cite{krishna_visual_2017}. We employ a top-$k$ strategy to sample the most probable nouns found in the video segment. We choose a \(k\) value of $5$ based on the experimental analysis by \citet{nam_zero-shot_2021}.
As for the CEM module, we rely on ConceptNet \cite{speer_conceptnet_2017} for commonsense information and extract the English sub-graph for our experiments. Moreover, we prune the commonsense graph \(G_{C}\) by preserving edge types that convey relevant contextual information, as detailed in Table 1 in the main paper.
We randomly initialize the weights for the GCN-based concept encoder.
Experiments for the balancing hyperparameter $\lambda$, spanning a range of $\lambda \in \{0.75, 0.7, 0.3, 0.25\}$ show that performance is consistently high across all metrics for $\lambda=0.7$, indicating the relative importance of temporal attention-guided loss over the overall localization regression loss.

\section{Ablation Studies}
\label{sec:ablations}
We further present ablation studies that focus on the Commonsense Enhancement Module (CEM) design and the overall efficacy of commonsense integration. Unless specified otherwise, we perform ablations on Charades-STA~\cite{Gao_2017_ICCV} and \modelname with 300 seed concepts.

\subsection{How to Best Inject Commonsense?}
\label{ablation:qcc}
\begin{figure}[t!]
    \centering
    \begin{subfigure}
        \centering
        \begin{tikzpicture}[scale=0.5]
          \begin{axis}[
            ybar,
            bar width=15pt,
            ymin=0,
            enlarge x limits={abs=25pt},
            legend style={draw=none,at={(0.5,-0.15)},
            anchor=north,legend columns=2},
            xlabel={Metric},
            ylabel={~~},
            nodes near coords,
            every node near coord/.append style={font=\normalsize,text width=0.5cm,rotate=90,align=center,
            xshift=-20pt,
            yshift=-6pt
            },
            symbolic x coords={$R@0.3$,$R@0.5$,$R@0.7$,$mIoU$}, 
            point meta=y,  
            xtick=data,
            legend to name={legqcc},
            legend image code/.code={%
                \draw[#1, draw=none] (0cm,-0.1cm) rectangle (0.6cm,0.1cm);
            },  
            legend style={
                draw=none, 
                text depth=0pt,
                at={(0.0,-0.15)},
                anchor=north west,
                legend columns=-1,
                column sep=5cm,
                /tikz/column 2/.style={column sep=15pt},
                %
                /tikz/every odd column/.append style={column sep=0cm},
            },
            cycle list={blueaccent,orangeaccent}
          ]	
            \addplot[fill=blueaccent] coordinates 
            {($mIoU$,32.73) ($R@0.3$,49.21) ($R@0.5$,34.60) ($R@0.7$,17.93)}; 
            \addplot[fill=orangeaccent] coordinates 
            {($mIoU$,30.33) ($R@0.3$,46.72) ($R@0.5$,29.33) ($R@0.7$,13.69)}; 
            \legend{Enhancement, Concatenation}
        \end{axis}
        \end{tikzpicture}
    \end{subfigure}%
    \begin{subfigure}
        \centering
        \begin{tikzpicture}[scale=0.5]
          \begin{axis}[
            ybar,
            bar width=15pt,
            ymin=0,
            enlarge x limits={abs=25pt},
            legend style={draw=none,at={(0.5,-0.15)},
            anchor=north,legend columns=2},
            xlabel={Metric},
            ylabel={~~},
            nodes near coords,
            every node near coord/.append style={
            font=\normalsize,
            text width=0.5cm,
            rotate=90,
            align=center,
            xshift=-20pt,
            yshift=-7pt},
            symbolic x coords={$R@0.3$,$R@0.5$,$R@0.7$,$mIoU$},
            point meta = y,  
            xtick=data,
          ] 			
            \addplot[fill=blueaccent] coordinates 
            {($mIoU$,33.06) ($R@0.3$,50.98) ($R@0.5$,33.18) ($R@0.7$,16.48)}; 
            \addplot[fill=orangeaccent] coordinates 
            {($mIoU$,29.38) ($R@0.3$,42.77) ($R@0.5$,31.40) ($R@0.7$,15.62)}; 
        \end{axis}
        \end{tikzpicture}
    \end{subfigure}
    \ref{legqcc}
    \vspace{-0.2cm}
    \caption{\modelname performance with enhancement \vs query-concepts concatenation for 300 (left) and 250 (right) seed concept sizes.}
\label{fig:ablation_qcc}
\end{figure}
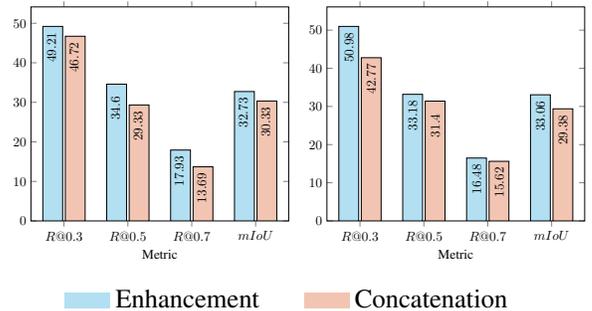

We evaluate the effectiveness of the proposed enhancement mechanism by comparing it against an alternate configuration that omits the enhancement process and instead concatenates the encoded concept vectors with the text query vectors, treating the resultant feature set as the text query features. Figure \ref{fig:ablation_qcc} shows the relative performance of concatenation \vs enhancement across both \modelname configurations with  300 or 250 seed concepts. We observe that enhancement consistently outperforms concatenation, thereby reinforcing the effectiveness of the enhancement flow in injecting necessary commonsense information. Notably, the concatenation configuration for 300 seed concepts still outperforms the PSVL baseline at various recall thresholds \ie, $k=\{0.3, 0.7\}$. This highlights the capacity of commonsense information to enhance localization performance even with a much simpler injection mechanism.

\subsection{Which Modality to Enhance with Commonsense?}
\label{ablation:cem}
To analyze the importance of commonsense enhancement across modalities, we train \modelname with the following configurations: (1) Only query features $Q$ are enhanced (\textbf{Q}), (2) Only video features $V$ are enhanced (\textbf{V}). In addition, we employ two configurations for which both video and query are enhanced. \modelname's CEM makes use of the same concept vectors, but employs separate enhancement steps for video and text query, \ie, we rely on two separate sub-modules $\phi_{C_{\text{vid}}}(V)$ and $\phi_{C_{\text{pq}}}(Q)$. This design choice stems from the hypothesis that the gap between video and query modalities is exacerbated when dealing with less sophisticated queries. Essentially, less sophisticated (or more general) queries may lack specificity or fail to capture the nuances of the desired information accurately. As a result, the gap between the information contained in the video and the intended query widens, making it more challenging to match the two modalities effectively. 
Having separate projection matrices for video and query allows differently enhancing $V$ and $Q$  with the same commonsense information (through the same concept vectors). 
To test this rationale, (3) we train \modelname with shared weights for $V$ and $Q$, \ie, $\phi_{C_{\text{vid}}}(V)$ and $\phi_{C_{\text{pq}}}(Q)$ are identical (\textbf{VQ}). Finally, (4) we represent the original setting of separate enhancement mechanisms for $V$ and $Q$ as \textbf{V+Q}.

\begin{table}[t!]
\centering
\resizebox{\linewidth}{!}{
\begin{tabular}{lcccc}
\toprule
\textbf{Method}            & \textbf{{R@0.3}}       & \textbf{{R@0.5}}       & \textbf{{R@0.7}}       & \textbf{{mIoU}}        \\\midrule
\modelname \textbf{{(Q)}}    & 39.72            & 22.16             & 8.10              & 26.06         \\
\modelname \textbf{{(V)}}    & {\underline{44.11}}  & {31.64}           & 15.20             & {\underline{29.75}}  \\
\modelname \textbf{{(VQ)}}   & 40.21            & \underline{31.78}     & \textbf{18.65}    & 28.45         \\
\modelname \textbf{{(V+Q)}}  & \textbf{49.21}   & \textbf{34.60}    & {\underline{17.93}}   & \textbf{32.73}       \\\bottomrule
\end{tabular}
}
\vspace{-0.2cm}
\caption{\modelname performance with query enhancement only (\textbf{{Q}}), video enhancement only (\textbf{{V}}), shared video and query enhancement (\textbf{{VQ}}) and separate video and query enhancement (\textbf{{V+Q}}). The best and second-best scores are shown in \textbf{bold} and \underline{underline}, respectively.}
\label{tab:weightsCEM}
\end{table}
Table \ref{tab:weightsCEM} presents results for the aforementioned configurations. We observe a significant drop in performance with \textbf{Q} across all metrics, which shows that the localization abilities are negatively impacted by omitting the video feature enhancement. To further support this observation, we see a consistent increase across all metrics for \textbf{V}, where only video features are enhanced and query feature enhancement is omitted. This highlights the positive impact of incorporating important commonsense information in the visual context for boosting model performance. 
Furthermore, we observe a consistent deterioration in model performance across all metrics in \textbf{VQ} except for $R@0.7$ when compared to \textbf{V+Q}. This could be attributed to the fact that a common enhancement flow for $V$ and $Q$ may potentially collapse diverging sources of information into one latent representation. Separating the enhancement for the two modalities allows disentangling the learned latent representations for video and pseudo-query, thereby capturing different relationships, but with the same underlying commonsense knowledge. Finally, \textbf{V+Q} performing the best across all the aforementioned configurations validates our hypothesis of maintaining separate enhancement flows for video and text query features.

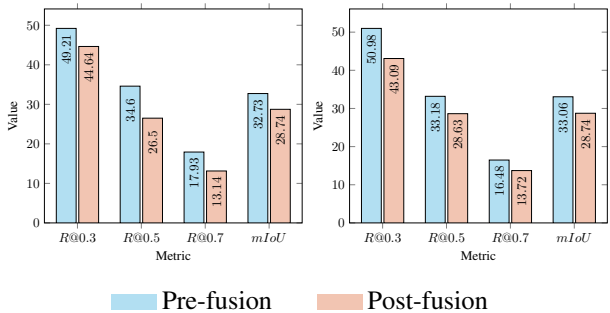
\begin{figure}[t!]
    \centering
    \begin{subfigure}
        \centering
        \begin{tikzpicture}[scale=0.5]
          \begin{axis}[
            ybar,
            bar width=15pt,
            ymin=0,
            enlarge x limits={abs=25pt},
            legend style={draw=none,at={(0.5,-0.15)},
            anchor=north,legend columns=2},
            xlabel={Metric},
            ylabel={Value},
            nodes near coords,
            every node near coord/.append style={font=\normalsize,text width=0.5cm,rotate=90,align=center,
            xshift=-20pt,
            yshift=-7pt
            },
            symbolic x coords={$R@0.3$,$R@0.5$,$R@0.7$,$mIoU$},
            point meta=y,  
            xtick=data,
            legend to name={legprepost},
            legend image code/.code={%
                \draw[#1, draw=none] (0cm,-0.1cm) rectangle (0.6cm,0.1cm);
            },  
            legend style={
                draw=none, 
                text depth=0pt,
                at={(0.0,-0.15)},
                anchor=north west,
                legend columns=-1,
                column sep=5cm,
                /tikz/column 2/.style={column sep=15pt},
                %
                /tikz/every odd column/.append style={column sep=0cm},
            },
            cycle list={blueaccent,orangeaccent}
          ]	
            \addplot[fill=blueaccent] coordinates 
            {($mIoU$, 32.73) ($R@0.3$, 49.21) ($R@0.5$,34.60) ($R@0.7$,17.93)}; 
            \addplot[fill=orangeaccent] coordinates 
            {($mIoU$,28.74) ($R@0.3$,44.64) ($R@0.5$,26.50) ($R@0.7$,13.14)}; 
            \legend{Pre-fusion,Post-fusion} 
        \end{axis}
        \end{tikzpicture}
    \end{subfigure}%
    \begin{subfigure}
        \centering
        \begin{tikzpicture}[scale=0.5]
          \begin{axis}[
            ybar,
            bar width=15pt,
            ymin=0,
            enlarge x limits={abs=25pt},
            legend style={draw=none,at={(0.5,-0.15)},
            anchor=north,legend columns=2},
            xlabel={Metric},
            ylabel={Value},
            nodes near coords,
            every node near coord/.append style={
            font=\normalsize,
            text width=0.5cm,
            rotate=90,
            align=center,
            xshift=-20pt,
            yshift=-7pt},
            symbolic x coords={$R@0.3$,$R@0.5$,$R@0.7$,$mIoU$},
            point meta = y,  
            xtick=data,
          ] 			
            \addplot[fill=blueaccent] coordinates 
            {($mIoU$,33.06) ($R@0.3$,50.98) ($R@0.5$,33.18) ($R@0.7$,16.48)}; 
            \addplot[fill=orangeaccent] coordinates 
            {($mIoU$,28.74) ($R@0.3$,43.09) ($R@0.5$,28.63) ($R@0.7$,13.72)}; 
        \end{axis}
        \end{tikzpicture}
    \end{subfigure}
    \ref{legprepost}
    \vspace{-0.2cm}
    \caption{\modelname performance with pre-fusion enhancement \vs post-fusion enhancement for 300 (left) and 250 (right) seed concept sizes.}
    \label{tab:ablation_pre_vs_post}
    \vspace{-0.3cm}
\end{figure}
\subsection{When to Perform Commonsense Enhancement?}
\label{ablation:prevpost}
\modelname separately enhances both video and query features prior to the cross-modal fusion step. However, an  alternative option would be to perform commonsense enhancement on the unified video-query features after cross-modal fusion and interaction. Accordingly, we present results for pre-fusion as well as post-fusion enhancement. 
In Figure \ref{tab:ablation_pre_vs_post}, we observe that our approach of pre-fusion enhancement works significantly better than post-fusion enhancement across both 300 and 250 concept sizes. We believe the underlying reason for this observation is consistent with our previous prior findings, where employing separate enhancement modules for video and query features is best suited to inject necessary information and allowed \modelname to differently approach video and query enhancement.
\subsection{Does Retaining Relational Information Boost Localization?}\label{ablation:relatinal_coronet}
\modelname employs a weighted directed graph as the concept graph $G_C$, where the edge weight between two nodes is the total number of relational edges from source to target nodes. In this ablation study, we analyze the impact of retaining multi-relational information in contrast to collapsing to a single weighted edge between two nodes. We replace the original $G_C$ version with a multi-relational directed graph, where each edge belongs to one of the relation types in Table 1 in the main paper, and two nodes may be connected with multiple different edges. To this end, we employ Relational Graph Convolutional Networks (RGCNs), where each graph convolution step is defined as
\begin{equation}
\label{eq:rgcn}
    C^{\left(l+1\right)}=\sigma\left(\sum_{r \in R} A_{r} C^{(l)} W_{r}^{(l)}\right).
\end{equation}
Here, $W_{r}^{(l)}$ is the trainable weight matrix for layer $l \in \left[1, L\right]$ and $A_r$ is the adjacency matrix for relation $r \in R$, where $R$ denotes our relation set. 
We experiment with both pre- and post-fusion enhancement in this setup and respectively denote them by \modelname-R and \modelname-R$_{post}$. 

\begin{table}[t!]
\centering
\resizebox{\linewidth}{!}{
\begin{tabular}{lcccc}
\toprule
\textbf{Model}  &   \textbf{R@0.3}  & \textbf{R@0.5}    &   \textbf{R@0.7}  & \textbf{mIoU}  \\ \midrule
\modelname      &   \textbf{49.21}         & \textbf{34.60}    &   \textbf{17.93}  & \textbf{{32.73}} \\
\modelname-R  & 40.52 & \underline{27.92} & \underline{13.85} & 27.80  \\
\modelname-R$_{post}$ & \underline{46.83} & 25.57 & 12.45 & \underline{30.91} \\ \bottomrule
\end{tabular}
}
\vspace{-0.2cm}
\caption{\modelname performance with multi-relational directed $G_C$ with pre- (\modelname-R) and post-fusion (\modelname-R$_{post}$) enhancement. The best and second-best scores are shown in \textbf{bold} and
\underline{underline}, respectively.}
\label{tab:ablation_relational}
\end{table}

Results in Table \ref{tab:ablation_relational} show that contrary to one's intuition, having a higher-order contextual graph with more relational information does not help localization performance. A multi-relational adjacency matrix is much sparser than its weighted counterpart, where the adjacency matrix is aggregated along the relation dimension. Having a denser adjacency matrix could possibly enhance learning. Overall, this experiment showcases that the association between two given objects is more important in integrating commonsense, rather than the specific relation type.
\subsection{Does Auxiliary Commonsense Information Boost Performance?}
\label{ablation:hops}
\begin{figure}[t!]
    \centering
    \begin{subfigure}
        \centering
        \begin{tikzpicture}[scale=0.5]
          \begin{axis}[
            ybar,
            bar width=15pt,
            ymin=0,
            enlarge x limits={abs=25pt},
            legend style={draw=none,at={(0.5,-0.15)},
            anchor=north,legend columns=2},
            xlabel={Metric},
            ylabel={Value},
            nodes near coords,
            every node near coord/.append style={font=\normalsize,text width=0.5cm,rotate=90,align=center,
            xshift=-20pt,
            yshift=-7pt},
            symbolic x coords={$R@0.3$,$R@0.5$,$R@0.7$,$mIoU$},
            point meta=y,  
            xtick=data,
            legend to name={leg},
            legend image code/.code={%
                \draw[#1, draw=none] (0cm,-0.1cm) rectangle (0.6cm,0.1cm);
            },  
            legend style={
                draw=none, 
                text depth=0pt,
                at={(0.0,-0.15)},
                anchor=north west,
                legend columns=-1,
                column sep=5cm,
                /tikz/column 2/.style={column sep=15pt},
                %
                /tikz/every odd column/.append style={column sep=0cm},
            },
            cycle list={blueaccent,orangeaccent}
          ]	
            \addplot[fill=blueaccent] coordinates 
            {($R@0.3$, 49.21) ($R@0.5$,34.60) ($R@0.7$, 17.93) ($mIoU$, 32.73)}; 
            \addplot[fill=orangeaccent] coordinates 
            {($R@0.3$, 42.29) ($R@0.5$,28.81) ($R@0.7$,14.68) ($mIoU$, 28.44)}; 
            \legend{0-Hop,1-Hop}
        \end{axis}
        \end{tikzpicture}
    \end{subfigure}%
    \begin{subfigure}
        \centering
        \begin{tikzpicture}[scale=0.5]
          \begin{axis}[
            ybar,
            bar width=15pt,
            ymin=0,
            enlarge x limits={abs=25pt},
            legend style={draw=none,at={(0.5,-0.15)},
            anchor=north,legend columns=2},
            xlabel={Metric},
            ylabel={Value},
            nodes near coords,
            every node near coord/.append style={
            font=\normalsize,
            text width=0.5cm,
            rotate=90,
            align=center,
            xshift=-20pt,
            yshift=-7pt},
            symbolic x coords={$R@0.3$,$R@0.5$,$R@0.7$,$mIoU$},
            point meta = y,  
            xtick=data,
          ] 			
            \addplot[fill=blueaccent] coordinates 
            {($mIoU$, 33.06) ($R@0.3$, 50.98) ($R@0.5$,33.18) ($R@0.7$,16.48)}; 
            \addplot[fill=orangeaccent] coordinates 
            {($mIoU$, 30.65) ($R@0.3$, 45.33) ($R@0.5$,30.99) ($R@0.7$,15.15)}; 
        \end{axis}
        \end{tikzpicture}
    \end{subfigure}
    \ref{leg}
    \vspace{-0.2cm}
    \caption{We compare \modelname performance with 1-hop neighborhood graphs with their 0-hop neighborhood graph counterparts for 300 (left) and 250 (right) seed concepts.}
    \label{fig:ablationHops}
    \vspace{-0.2cm}
\end{figure}
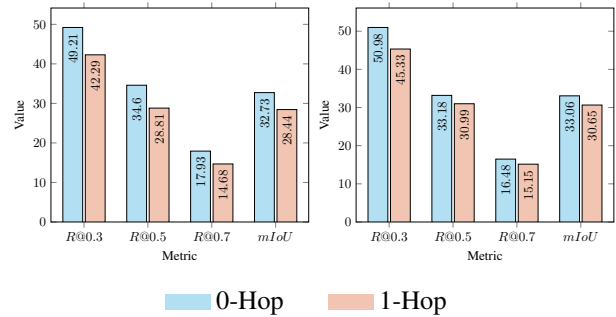
We analyze the impact of including auxiliary contextual information provided through \(G_{C}\). We examine the performance of \modelname by replacing the proposed seed concept graph \(G_{C}\) with a bigger $1$-hop neighborhood graph. Since including a 1-hop neighborhood leads to an exponential increase in the graph size, we limit $G_C$ to include 1-hop neighborhood only with edge types that add valuable information to the localization setup. Specifically, we include edge types that may involve action information (\ie, verbs) in relation to the objects observed in our video corpus (\eg, $UsedFor$, $CapableOf$, $Causes$, \etc). 
Figure~\ref{fig:ablationHops}
shows the relative performance of this model variant in comparison to the original seed concept (0-hop) graph. Performance consistently worsens across all metrics for both 300 and 250 seed concept sizes, with more drastic drops for 300 concept sizes. We hypothesize that, despite the increased context via a larger graph, the additional information may prove to be noisy, thereby affecting localization accuracy as well as generalization capabilities.
\begin{table}[t!]
\centering
\resizebox{\columnwidth}{!}{
\begin{tabular}{lccccc}
\toprule
{Encoder} & \textbf{R@0.3} & \textbf{R@0.5} & \textbf{R@0.7} & \textbf{mIoU}  & \textbf{time/epoch} \\ \midrule
{GRU}        	&49.21	&{\textbf{34.60}}&	{\textbf{17.93}} & {\textbf{32.73}} & 74.48s                  \\
{Transformer}  	&{\textbf{53.57}}	&30.67	&13.49  &32.70        & 35.94s                  \\ \bottomrule
\end{tabular}}
\vspace{-0.2cm}
\caption{Performance with recurrent \vs Transformer-based encoders for video and query inputs. Time per epoch is measured in seconds. The best scores are presented in \textbf{bold}.}
\label{tab:encoderAblation}
\end{table}
\subsection{How to Best Encode Inputs?}
\label{ablation:encoder}
We also investigate the impact of adopting a recurrent architecture (GRU/LSTM) \vs Transformers~\cite{vaswani_attention_2017} for generating the video $V$ and pseudo-query $Q$ encodings. Table \ref{tab:encoderAblation} quantitatively compares model performance under such encoding variants for \modelname.
While Transformer-based methods are more than twice as fast as recurrent methods, they surprisingly impede model performance by large margins across most metrics.

\subsection{Does Commonsense Help in Language-free Setups?}
\label{ablation:lfvl}
\begin{table}[t!]
\centering
\resizebox{\columnwidth}{!}{
\begin{tabular}{lccccc}
\toprule
\textbf{Model} & \textbf{Enhancement} & \textbf{R@0.3} & \textbf{R@0.5} & \textbf{R@0.7} & \textbf{mIou} \\ \midrule
LFVL~\cite{kim2023language}           & None                          & 49.50 & \textbf{34.39} & \textbf{16.95} & 33.19         \\ \midrule
+ CEM          & \multirow{2}{*}{Post}         & \textbf{54.39} & 31.38 & \underline{14.29} & \underline{34.19}         \\
+ CEM$_{250}$  &                               & \underline{53.26} & \underline{33.05} & 13.99 & \textbf{34.30}         \\ \midrule
+ CEM          & \multirow{2}{*}{Pre}          & 49.01 & 28.98 & 12.97 & 31.30         \\
+ CEM$_{250}$  &                               & 49.16 & 29.56 & 13.36 & 31.67         \\ \bottomrule
\end{tabular}}
\vspace{-0.2cm}
\caption{Commonsense enhancement integrated to the LFVL~\cite{kim2023language} method. CEM and CEM$_{250}$ represent enhancement using our commonsense enhancement module with 300 and 250 seed concept graphs ($G_{C}$). Results for both post- and pre-fusion enhancement across CEM and CEM$_{250}$ are displayed. The best and second-best scores are shown in \textbf{bold} and \underline{underline}, respectively.}
\label{tab:lfvl}
\vspace{-0.3cm}
\end{table}

We also conduct an experiment to test the effectiveness of our CEM approach on a language-free NLVL (LFVL) setting \cite{kim2023language}. LFVL eliminates the need for query annotations by leveraging the cross-modal understanding of CLIP~\cite{radford2021learning} to utilize visual features as textual information. We integrate our commonsense enhancement mechanism into the LFVL pipeline to analyze its impact in this setup. Table \ref{tab:lfvl} compares model performances with two variants, CEM and CEM$_{250}$, which respectively contain 300 and 250 seed concepts. Furthermore, we examine the effectiveness of commonsense enhancement in a post- and pre-fusion setup.
We find that there is a significant increase in the $mIoU$ and $R@0.3$ scores with both CEM and CEM$_{250}$ in post-fusion setup. 
This indicates that the integration of commonsense enhancement positively impacts the overall localization performance. Notably, the comparison between post- and pre-fusion enhancement reveals a striking difference in performance. These findings suggest that enhancing the fused video-query representation with commonsense information is more beneficial compared to enhancing a language-free query representation. The results imply that enhancing the fused representation allows for a more effective alignment between video and query, leading to improved localization performance.

\end{document}